\documentclass[10pt,twocolumn,letterpaper]{article}

\usepackage{cvpr}
\usepackage{times}
\usepackage{epsfig}
\usepackage{graphicx}
\usepackage{amsmath}
\usepackage{amssymb}

\usepackage{algorithm}
\usepackage{algorithmic}
\usepackage{xcolor}
\usepackage{multirow}
\usepackage{subfig}
\usepackage{amsfonts}       

\usepackage[pagebackref=true,breaklinks=true,letterpaper=true,colorlinks,bookmarks=false]{hyperref}

\cvprfinalcopy 

\def\cvprPaperID{9817} 

\ifcvprfinal\fi

\newcommand{\name}{PhysGAN}
\newcommand{\advobject}{roadside sign}
\newcommand{\advobjects}{roadside signs}

\newcommand{\E}{encoder $\mathcal{E}$}
\newcommand{\G}{generator $\mathcal{G}$}
\newcommand{\D}[1][discriminator]{{#1} $\mathcal{D}$}
\newcommand{\T}[1][target autonomous driving model]{{#1} $f$}

\newcommand{\sadv}{S_{adv}}
\newcommand{\sorig}{S_{orig}}
\newcommand{\xadv}{X_{adv}}
\newcommand{\xorig}{X_{orig}}

\newcommand{\origv}[1][original video slice]{{#1} $\xorig{}$}
\newcommand{\advv}[1][adversarial video slice]{{#1} $\xadv{}$}
\newcommand{\advs}[1][adversarial roadside sign]{{#1} $\sadv{}$}
\newcommand{\origs}[1][original roadside sign]{{#1} $\sorig{}$}

\newcommand{\lgan}{\mathcal{L}_{GAN}}
\newcommand{\ladv}{\mathcal{L}^f_{ADV}}

\begin{document}

\title{\name: Generating Physical-World-Resilient Adversarial Examples \\ for Autonomous Driving}

\author{Zelun Kong, Junfeng Guo, Ang Li,~and Cong Liu\\
The University of Texas at Dallas \\
{\tt\small \{zelun.kong,junfeng.guo,angli,cong\}@utdallas.edu}
}

\maketitle 

\begin{abstract}
    Although Deep neural networks (DNNs) are being pervasively used in vision-based autonomous driving systems, they are found vulnerable to adversarial attacks where small-magnitude perturbations into the inputs during test time cause dramatic changes to the outputs. 
    While most of the recent attack methods target at digital-world adversarial scenarios, it is unclear how they perform in the physical world, and more importantly, the generated perturbations under such methods would cover a whole driving scene including those fixed background imagery such as the sky, making them inapplicable to physical world implementation. 
    We present \name{}, which generates physical-world-resilient adversarial examples for misleading autonomous driving systems in a continuous manner. 
    We show the effectiveness and robustness of \name{} via extensive digital- and real-world evaluations. 
    We compare \name{} with a set of state-of-the-art baseline methods, which further demonstrate the robustness and efficacy of our approach. We also show that \name{} outperforms state-of-the-art baseline methods. To the best of our knowledge, \name{} is probably the first technique of generating realistic and physical-world-resilient adversarial examples for attacking common autonomous driving scenarios.
\end{abstract}

\section{Introduction}

\begin{figure}
    \centering
        \includegraphics[width=.49\columnwidth,height=72px]{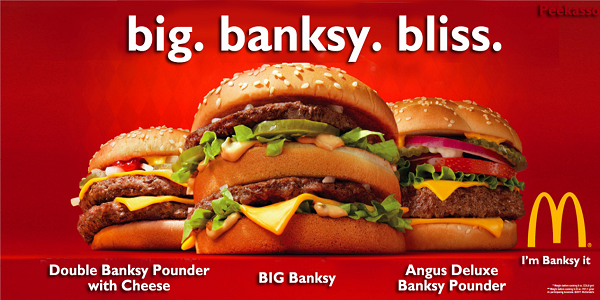}
        \includegraphics[width=.49\columnwidth,height=72px]{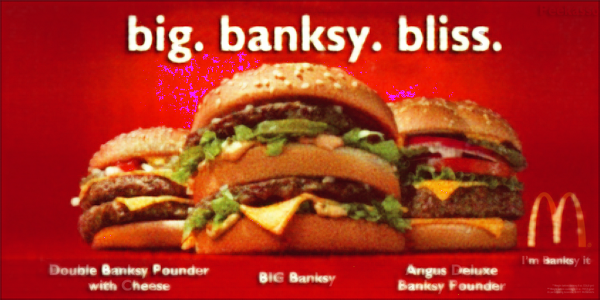}
        \includegraphics[width=.99\columnwidth,height=72px]{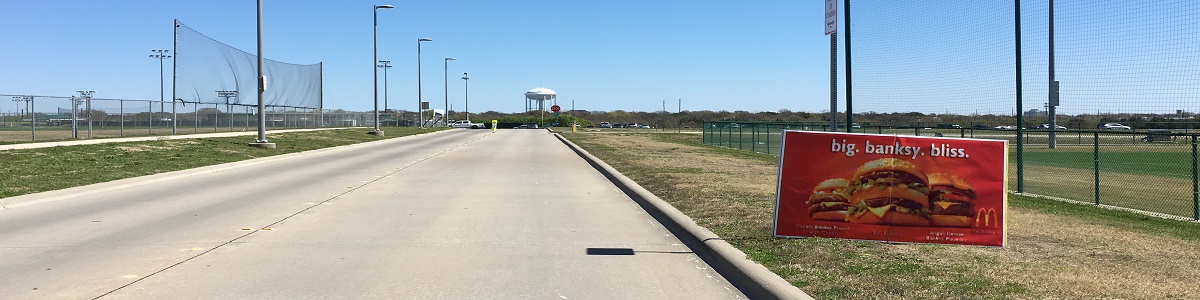}
    \caption{\small Illustration of an adversarial roadside advertising sign (top-right) visually indistinguishable from the original sign  (top-left) and its deployment in the physical world (bottom).}
    \vspace{-6mm}
    \label{fig:introeg}
\end{figure}

While deep neural networks (DNNs) have established the fundamentals of vision-based autonomous driving systems, they are still vulnerable to adversarial attacks and exhibit erroneous fatal behaviors. 
Recent works on adversarial machine learning research have shown that DNNs are rather vulnerable to intentional adversarial inputs with perturbations focusing on classification problems~\cite{carlini2017,kurakin2016,moosavi2016,papernot2016,szegedy2013}. To address the safety issues in autonomous driving systems, techniques were proposed to automatically generate adversarial examples, which add small-magnitude perturbations to inputs to evaluate the robustness of DNN-based autonomous driving systems~\cite{kurakin2016, eykholt2018, tian2018}.

However, these techniques mostly focus on generating \emph{digital} adversarial examples, \eg, changing image pixels, which can never happen in real world~\cite{kurakin2016}. They may not be applicable to realistic driving scenarios, as the perturbations generated under such techniques would cover the whole scene including fixed background imagery such as the sky. Very recently, a few works took the first step in studying physical-world attacking/testing of static physical objects~\cite{athalye2017,metzen2017}, human objects~\cite{sharif2016,elsayed2018}, traffic signs~\cite{sermanet2011,mogelmose2012, eykholt2018}.
Although they are shown effective under the targeted scenarios and certain assumptions, they focus on studying a static physical-world scene (\eg, a single snapshot of a stop sign~\cite{eykholt2018,sermanet2011}), which prevent themselves to be applied in practice as real-world driving is a continuous process where dynamically changes are encountered (\eg, viewing angles and distances).
Moreover, their generated adversarial examples are visually unrealistic (\eg, driver-noticeable black and white stickers pasted onto a stop sign which is easily noticeable for attack purposes~\cite{eykholt2018}). Most of these methods also have a focus on classification models different from our studied steering model which is a regression model. 
Also note that directly extending the existing digital perturbation generation techniques (e.g., FGSM) to physical world settings, i.e., inputting just the targeted roadside sign into such techniques would output a corresponding adversarial example, may be ineffective. The resulting attack efficacy may dramatically decrease (proved in our evaluation as well) as the process of generating perturbations has not considered any potential background imagery in the physical world (e.g., the sky) which would be captured by any camera during driving.

We aim at generating a realistic single adversarial example which can be physically printed out to replace the corresponding original roadside object, as illustrated in Fig.~\ref{fig:introeg}. 
Since the target vehicle observes this adversarial printout continuously, a major challenge herein is how to generate a single adversarial example which can continuously mislead the steering model at every frame during the driving process. Additionally, for a practical physical-world deployment, any generated adversarial example shall be visually indistinguishable from its original sign (the one already deployed in the physical world). 

To address these challenges, we propose a novel GAN-based framework called \name{}~\footnote{https://github.com/kongzelun/physgan.git} which generates a single adversarial example by observing multiple frames captured during driving while preserving resilience to certain physical-world conditions.
Our architecture contains an encoder (\ie, the CNN part of the target autonomous driving model) that extracts features from frames during driving and transforms them into a vector serving as the input to the generator. By considering all factors extracted from the frames, this design ensures that the generator could generate adeversarial examples that have attack effect.
Without this encoder, the efficacy would dramatically decrease.
To generate an adversarial example that can continuously mislead the steering model, \name{} takes a 3D tensor as input. This enhances the resilience of the generated example against certain physical world dynamics, as using video slices makes it more likely to capture such dynamics.

We demonstrate the effectiveness and robustness of \name{} through conducting extensive digital- and real-world experiments using a set of state-of-the-art steering models and datasets. Digital experimental results show that \name{} is effective for various steering models and scenarios, being able to mislead the average steering angle by up to 21.85 degrees. Physical case studies further demonstrate that \name{} is sufficiently resilient in generating physical-world adversarial examples, which is able to mislead the average steering angle by up to 19.17 degrees. Such efficacy is also demonstrated through comparisons against a comprehensive set of baseline methods.


To the best of our knowledge, \name{} is the first technique of generating realistic and physical-world-resilient adversarial examples for attacking common autonomous steering systems. Our contributions can be summarized in three folds as follows.
\begin{itemize}
\itemsep 0pt
    \item We propose a novel GAN-based framework \name{} which can generate physical-world-resilient adversarial examples corresponding to any roadside traffic/advertising sign and mislead autonomous driving steering model with the generated visually indistinguishable adversarial examples.  
    \item We propose a GAN architecture using 3D tensor as input in optimizing the generator, which resolves a key technical challenge in physical-world deployment of using a single adversarial example to continuously mislead steering during the entire driving process.
    \item We conduct extensive digital and physical-world evaluations with several metrics, which shows the superior attack performance of PhysGAN over state-of-the-art methods. We believe PhysGAN could contribute to future safety research in autonomous driving.
\end{itemize}

\section{Related Works}
\textbf{Adversarial Attacks.}
Many works have recently been proposed to generate adversarial examples for attacking in the white-box setting~\cite{papernot2017}, where the adversary knows the network's parameters. The fast gradient sign method (FGSM)~\cite{Goodfellow} represents the pioneer among such methods, which performs a one-step gradient update along the direction of the sign of gradient at each pixel. 
 FGSM is further extended in \cite{kurakin2017} to a targeted attack strategy through maximizing the probability of the target class, which is referred as the OTCM attack. Optimization-based approaches~\cite{tanay2016boundary,liu2016delving,carlini2017,carlini2017b,xiao2018} have also been proposed. GAN was recently introduced in \cite{GAN}, implemented by a system of two neural networks competing with each other in a zero-sum game framework. GAN achieves visually appealing results in both face generation~\cite{lu2017conditional} and manipulation~\cite{zhu2016generative}. \cite{xiao2018} presents AdvGAN, which leverages GAN to produce adversarial examples with high attack success rate on classification problems.
These methods focus on applying perturbations to the entire input and consider only digital-world attacking scenarios. It is hard to apply them to the real world because it is impossible to use some of the generated perturbations to replace the real-world background (\eg, the sky).

\textbf{Generating Physical  Adversarial Examples.}
To the best of our knowledge, only a very recent set of works~\cite{lu2017no,eykholt2018} started working on generating physical attacks. \cite{lu2017no} focuses on the understanding of static physical adversarial examples. \cite{eykholt2018} explicitly designs perturbations to be effective in the presence of diverse real-world conditions. Their method mainly focuses on the classification of physical road sign under dynamic distance and angle of the viewing camera. Unfortunately, these works focus on static attacking scenarios (e.g., maximizing the adversarial effectiveness w.r.t. to a single snapshot of the physical example) and thus do not require to resolve the one-to-many challenge. 

Different from these works, \name{} is able to generate physical-world-resilient adversarial examples only corresponding to the roadside traffic/advertising sign; no perturbations will be generated on areas other than the street sign. \name{} addresses the one-to-many challenge which continuously attack the steering model, and generates realistic adversarial examples that are resilient to various physical world conditions and visually indistinguishable from the original roadside sign.

\section{Our Approach: \name}

\begin{figure}[t]
    \centering
    \includegraphics[width=1.0\columnwidth]{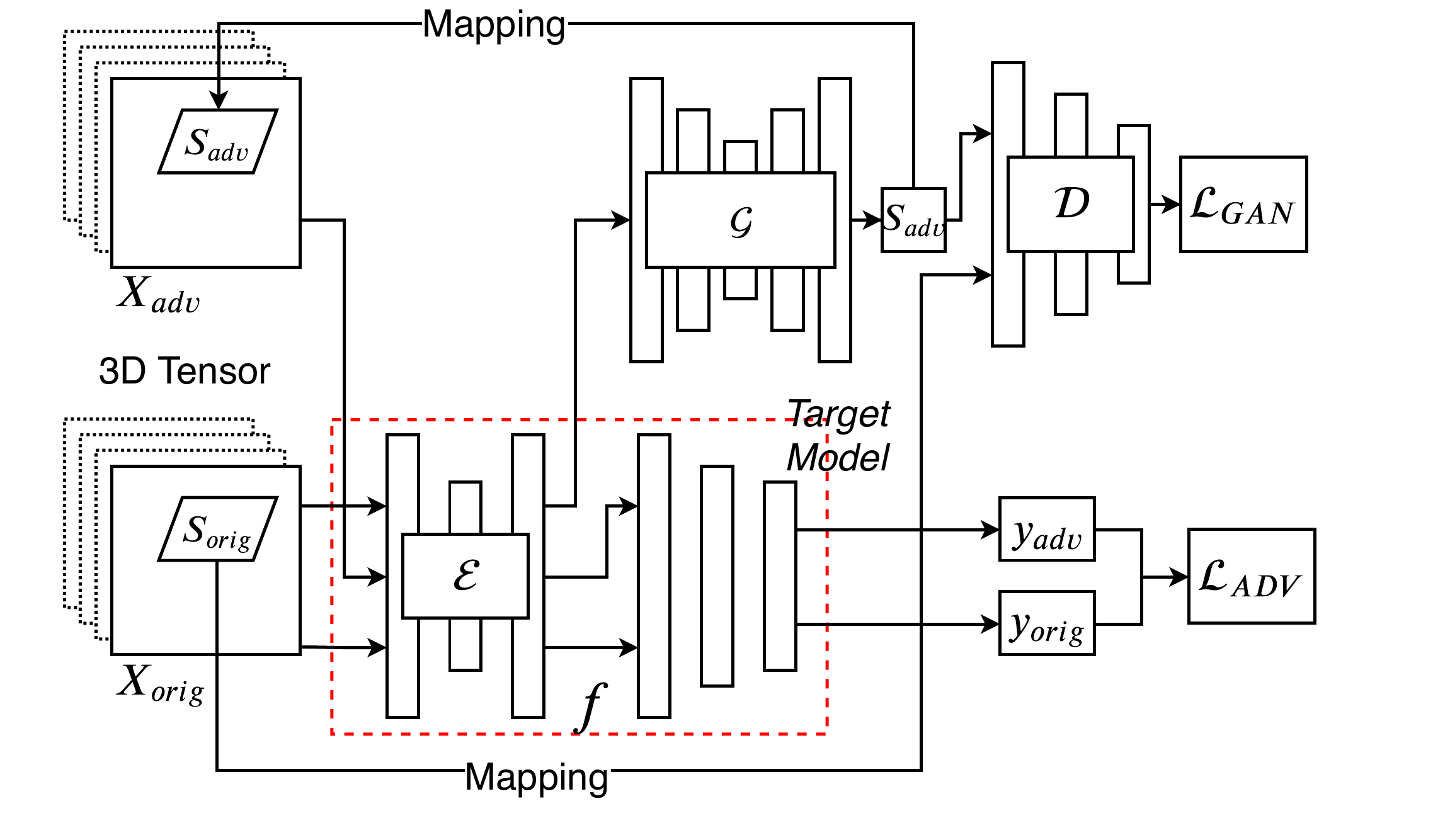}
    \caption{Overview of the \name{} framework.}
    \label{fig:framework}
\end{figure}

The goal of \name{} is to generate an adversarial example that is visually indistinguishable from any common roadside object (\eg, roadside traffic or advertising signs) to continuously mislead the steering angle model (target model) of a drive-by autonomous driving vehicle by physically replacing the roadside board with the adversarial example. When an autonomous driving vehicle drives by the \advobject, the steering angle model would be fooled and make incorrect decision. 

\subsection{Problem Definition}
We define our problem and notations in this section.
Let $\mathcal{X}=\{X_i\}$ be the video slice set such that $\mathcal{X} \subseteq \mathbb{R}^{n \times w \times h}$, where $n$ is the number of frames in the video slice, $w$ and $h$ is the \textit{width} and \textit{height} of a frame, respectively.
Let $\mathcal{Y}=\{Y_i\}$ be the ground truth steering angle set, $\mathcal{Y} \subseteq \mathbb{R}^n$. 
Suppose $(X_i, Y_i)$ is the $i^{th}$ sample in the dataset, which is composed of video slice $X_i \in \mathcal{X}$ and $Y_i \in \mathcal{Y}$, each element of which denotes the ground truth steering angle corresponding to its frame.
The pre-trained target steering model (e.g., Nvidia Dave-2, Udacity Cg23 and Rambo) learns a mapping $f \colon \mathcal{X} \to \mathcal{Y}$ from the video slice set $\mathcal{X}$ to the ground truth steering angle set $\mathcal{Y}$ during training phase.

Given an instance $(X_i, Y_i)$, the goal of \name{} is to produce an \advs{}, which aims to mislead the \T{} as $f(X_i) \neq Y_i$ and maximize $\sum_i |f(X_i) - Y_i|$.  
To achieve the goal, \name{} needs to generate an \advs{} to replace \origs{} in digital- or physical-world.
The \advs{} is supposed to be close to the \origs{} in terms of $\ell^2$-norm distance metrics, which implies that \advs{} and \origs{} are almost visually indistinguishable.

\subsection{Physical World Challenges}
Physical attacks on an object must be able to survive
changing conditions and remain effective at fooling the steering angle model. We structure our decision of these conditions around the common drive-by scenarios (\ie, the vehicle drives towards the \advobject). 

\textbf{The ``One-to-Many'' challenge.} A key technical challenge is to resolve the ``one-to-many'' challenge, \ie, generating a single adversarial sample to continuously mislead the steering angle decision of a vehicle throughout the entire driving process. Considering multiple frames in generating an adversarial sample is challenging because the vehicle-to-board distance, view angle, and even subtle pixels on each frame could be different. An effective adversarial sample must be able to exhibit maximum overall attack effect among all the frames. To achieve this goal, the adversarial sample needs to be resilient to the changing conditions exhibited on each frame. 
To resolve this problem, \name{} applies a novel GAN-based framework and consider the entire drive-by video slice, rather than a single frame, as the input in the generation process (see Sec.~\ref{sec:attacks}). 

\textbf{Limited manipulated area.} Unlike most digital-world adversarial methods which add perturbations to the entire input image, techniques focused on physical-world scenarios are constrained to add perturbations only to a fragment of an image, \ie, the fragmented area corresponding to the original physical object. Moreover, the underlying assumption of a static image background does not hold in physical attacks since the background can consistently change over the driving process.

\subsection{\name{} Overview}
Fig.~\ref{fig:framework} illustrates the overall architecture of \name{}, which mainly consists of four components: an \E{}, a \G{}, a \D{} and the \T{}. 
The \E{} represents the convolutional layers of \T{}, which takes 3D tensors as inputs and is used to extract features of a video (of both original and perturbed ones). 
To resolve the challenge of generating only a single example which continuously impacts the driving process, we introduce a novel idea of considering 3D tensors as inputs in the GAN-based framework. 
2D tensors often represent images while 3D tensors are used to represent a small slice of video, which typically contains hundreds of frames. 

As seen in Fig.~\ref{fig:framework}, the extracted features of \origv{} are used as the input fed to the generator to generate an \advs{}. 
Doing so allows us to take into account the fact that different \origv{} may have different influence on the generated \advs{}, thus ensuring the \G{} to generate the best \advs{} corresponding to a certain \origv{}.  The \advs{} and \origs{} are sent to the \D{}, which is used to distinguish the \advs{} and the \origs{}. 
The \D{} represents a loss function, which measures the visual distinction between \advs{} and \origs{}, and also encourages the generator to generate an  example visually indistinguishable to the original sign.

\subsection{Training GAN Along With the Target Model}
In order to ensure the \advs{} to have adversarial effect on the \T{}, we introduce the following loss function:
\begin{equation}
    \ladv = \beta \exp\left(-\frac{1}{\beta} \cdot l_f (f(\xorig{}), f(\xadv{}))\right)
    \label{eq:adv_loss}
\end{equation}
where $\beta$ is a sharpness parameter, $l_f$ denotes the loss function used to train the \T{}, such as MSE-loss or $\ell^1$-loss, $\xorig$ denotes the \origv, and $\xadv$ represents the \advv, which is generated by mapping the \advs{} into every frame of the \origv. 
By minimizing $\ladv$, the distance between the prediction and the ground truth will be maximized, which ensures the adversarial effectiveness.


To compute $\ladv$, we obtain the \advv{} by substituting the \origs{} with the generated \advs{}. 
Note that the generated \advs{} is a rectangular image and the \origs{} in the video slice may exhibit an arbitrary quadrilateral shape which could vary among different frames.
We leverage a classical perspective mapping method~\cite{persepective_mapping}
to resolve this mismatch. We first get the four coordinates of the \origs{} within each frame, and then map the generated \advs{} onto the corresponding quadrilateral area inside each frame (details can be found in supplementary material).

The final objective of \name{} is expressed as:
\begin{equation}
    \mathcal{L} = \lgan + \lambda \ladv,
    \label{eq:full_loss}
\end{equation}
where $\lambda$ denotes a co-efficient to balance the tradeoff between the two terms and $\lgan$ is the classic GAN loss, which can be represented as
\begin{equation}
    \begin{aligned}
        \lgan
        & = \mathbb{E}_{{\sorig} \sim p_{\sorig}}  \left[\log \mathcal{D}(\sorig)\right] \\
        & + \mathbb{E}_{\sadv \sim p_{\sadv}} \left[\log (1 - \mathcal{D}(\sadv))\right]~.
    \end{aligned}
    \label{eq:gan_loss}
\end{equation}
To interpret this objective function, $\lgan$ encourages the \advs{} to be visually similar to the \origs{}, while $\ladv$ is leveraged to generate \advv{} which maximizes attack effectiveness. We obtain the \E{}, the \G{}, and the \D{} by solving:
\begin{equation}
    \arg \min_{\mathcal{G}} \max_{\mathcal{D}} \mathcal{L}.
\end{equation}

\subsection{Attacks with \name}
\label{sec:attacks}
\begin{algorithm}[!t]
    \begin{algorithmic}[1]
        \REQUIRE $I$ - Iteration numbers;
        \REQUIRE $f$ - Target model with fixed parameters;
        \WHILE{$i < I$:}
        \STATE $S_{adv} = \mathcal{G}(\mathcal{E}(\xorig{}))$;
        \STATE $\lgan = \log \mathcal{D}(\sorig{}) + \log (1 - \mathcal{D}(\sadv{}))$;
        \STATE // fix the parameters of $\mathcal{G}$
        \STATE do back-propagation to optimize $\arg \max_{\mathcal{D}} \lgan$;
        \STATE $S_{adv} = \mathcal{G}(\mathcal{E}(\xorig{}))$
        \STATE $\lgan = \log \mathcal{D}(\sorig{}) + \log (1 - \mathcal{D}(\sadv{}))$;
        \STATE // fix the parameters of $\mathcal{D}$
        \STATE for each frame in the input video slice, perform perspective mapping to substitute the \origs{} using the \advs{}. 
        \STATE do back-propagation to optimize $\arg \min_{\mathcal{G}} \lgan$;
        \STATE $\mathcal{L}_{ADV} = \beta \exp(-\frac{1}{\beta} \cdot l_f (f(\xorig{})))$;
        \STATE do back-propagation to optimize $\arg \min_{\mathcal{G}} \mathcal{L}_{ADV}$;
        \ENDWHILE
    \end{algorithmic}
    \caption{\textbf{Optimization for \name{}}}
    \label{alg:optimize}
\end{algorithm}

We assume that the \T{} was pre-trained and the parameters of \T{} are fixed, and the \G{} of \name{} can only access the parameters of the \T{} during training. Our algorithm to train \name{} is illustrated in Algorithm~\ref{alg:optimize}, which consists of two phases.  
As seen in Algorithm~\ref{alg:optimize}, the first phase is to train the \D{}, which is used later to form a part of the $\lgan$ (Line 2 -- 5); the second phase is to train \G{} with two loss functions, $\ladv$ and $\lgan$, which encourages the \G{} to generate a visually indistinguishable adversarial sample and make the generated sample be adversarial for the \T{}, respectively~(Line 6 -- 11). 
The \E{}, which is the CNN part of the \T{}, aims at extracting features from all the observed frames during driving and transforms them into a vector input to the generator. This design ensures that the generator could generate visually indistinguishable examples with an attack effect through considering all useful features extracted from the video slice.
For physical world deployment, the attacker shall print the adversarial example of the same size as the target roadside sign to ensure visual indistinguishability.

\section{Experiments}

We evaluate \name{} via both digital and physical-world evaluations using widely-studied CNN-based steering models and popular datasets. 


\subsection{Experiment Setup}
\textbf{Steering models. }
We evaluate \name{} on several popular and widely-studied~\cite{chen2015,tian2018deeptest,broggi2013} CNN-based steering models, like NVIDIA Dave-2~\footnote{https://devblogs.nvidia.com/deep-learning-self-driving-cars/}, Udacity Cg23~\footnote{https://github.com/udacity/self-driving-car/tree/master/steering-models/community-models/cg23} and Udacity Rambo~\footnote{https://github.com/udacity/self-driving-car/tree/master/steering-models/community-models/rambo}. 
Notably, since the original models applies 2D CNNs which is trained with individual images, we adapt the 2D CNN into a 3D CNN, and train the 3D-CNNs with a set of 20-frame video slices.

\textbf{Datasets.} The datasets used in our digital experiments include 
(1) Udacity automatic driving car challenge dataset~\footnote{https://medium.com/udacity/challenge-2-using-deep-learning-to-predict-steering-angles-f42004a36ff3}, which contains 101396 training images captured by a dashboard mounted camera of a driving car and the simultaneous steering wheel angle applied by a human driver for each image; 
(2) DAVE-2 testing dataset~\cite{pan2017virtual}~\footnote{https://github.com/SullyChen/driving-datasets}, which contains 45,568 images to test the NVIDIA DAVE-2 model; 
(3) Kitti \cite{geiger2013vision} dataset which contains 14,999 images from six different scenes captured by a VW Passat station wagon equipped with four video cameras; and
(4) custom datasets for physical-world evaluation, which contain more than 20000 frames used to train \name{} in physical cases.

For physical-world experiments, we first perform color augmentation to improve the image contrast, making the adversarial example be more robust against varying light illumination conditions. 
Then, we print out the generated example under each evaluated approach, and paste it on the selected roadside object. We drive a vehicle by this object and perform offline analysis using the captured driving videos. To understand how \name{} would perform on actual autonomous vehicles, we have also done online driving testing which mimics a realistic driving process when facing with such an adversarial roadside object.

\begin{table}[h]
    \centering
    \resizebox{1.0\columnwidth}{!}{
    \begin{tabular}{|l|c|c|c|c|}
        \hline
        Scenes & Images & Size & min & max \\
        \hline\hline
        Dave-straight1 & 20 & $455 \times 256$ & $21 \times 22$ & $41 \times 49$ \\
        \hline
        Dave-curve1 & 20 & $455 \times 256$ & $29 \times 32$ & $51 \times 49$ \\
        \hline
        Udacity-straight1 & 20 & $640 \times 480$ & $48 \times 29$ & $66 \times 35$ \\
        \hline
        Udacity-curve1 & 20 & $640 \times 480$ & $51 \times 51$ & $155 \times 156$ \\
        \hline
        Kitti-straight1 & 20 & $455 \times 1392$ & $56 \times 74$ & $121 \times 162$ \\
        \hline
        Kitti-straight2 & 20 & $455 \times 1392$ & $80 \times 46$ & $247 \times 100$ \\
        \hline
        Kitti-curve1 & 20 & $455 \times 1392$ & $64 \times 74$ & $173 \times 223$ \\
        \hline
    \end{tabular}
    }
    \caption{Scenes evaluated in the experiment.}
    \label{tab:scenes}
\end{table}

\textbf{Video slice selection criteria.}
Our driving scene selection criteria is that the roadside traffic or advertising signs should appear entirely in the first frame of a driving video with more than 400 pixels and partially disappear in the last frame. We select 7 scenes from the aforementioned datasets, and evaluate on all selected scenes. The selected scenes in each dataset cover both straight and curved lane scenarios. Since all these datasets do not contain coordinates of roadside signs, we have to label the four corners of the signs in every frame of the selected scenes.
We use the motion tracker functionality of Adobe After Effects~\footnote{https://www.adobe.com/products/aftereffects.html} to automatically track the movement of the sign’s four corners among consecutive frames. 
Table~\ref{tab:scenes} show the attributes of the scenes we selected.


\newcommand{\includefigure}[1]{\hspace{-5.5pt}\begin{minipage}{62pt}\includegraphics[width=68pt,height=51pt]{#1}\end{minipage}}

\begin{table*}[t]
    \centering
    \begin{tabular}{|c|c|c|c|c|c|c|c|}
        \hline
        \multirow{2}{*}{} & \multicolumn{2}{c|}{Dave} & \multicolumn{2}{c|}{Udacity} & \multicolumn{3}{c|}{Kitti} \\
        \cline{2-8} & Straight1 & Curve1 & Straight1 & Curve1 & Straight1 & Straight2 & Curve1 \\
        \hline
        \rotatebox[origin=c]{90}{Apple} &
        \includefigure{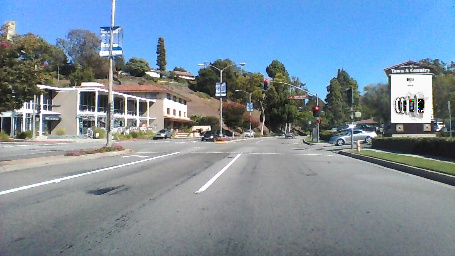} &
        \includefigure{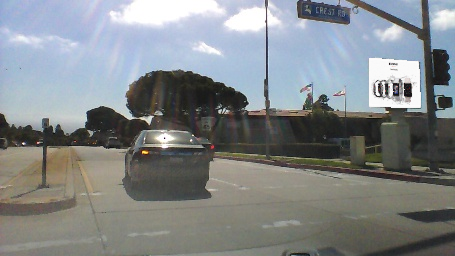} &
        \includefigure{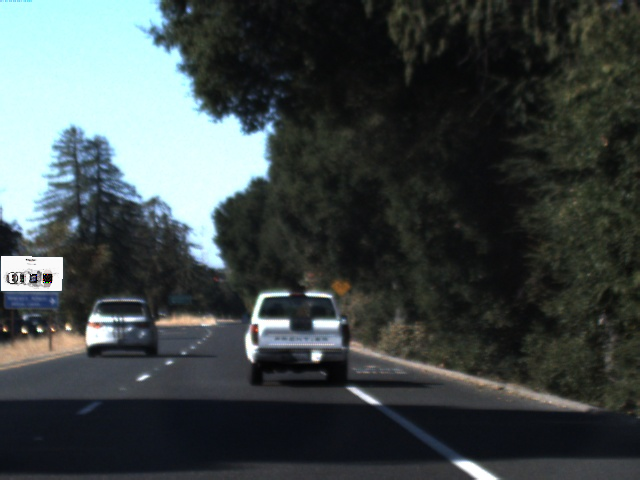} &
        \includefigure{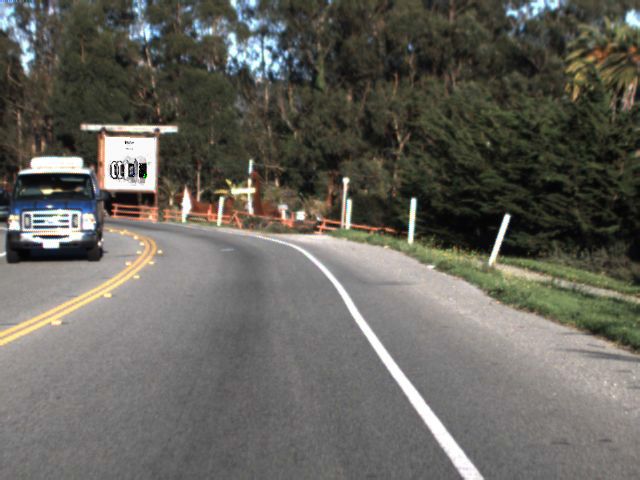} &
        \includefigure{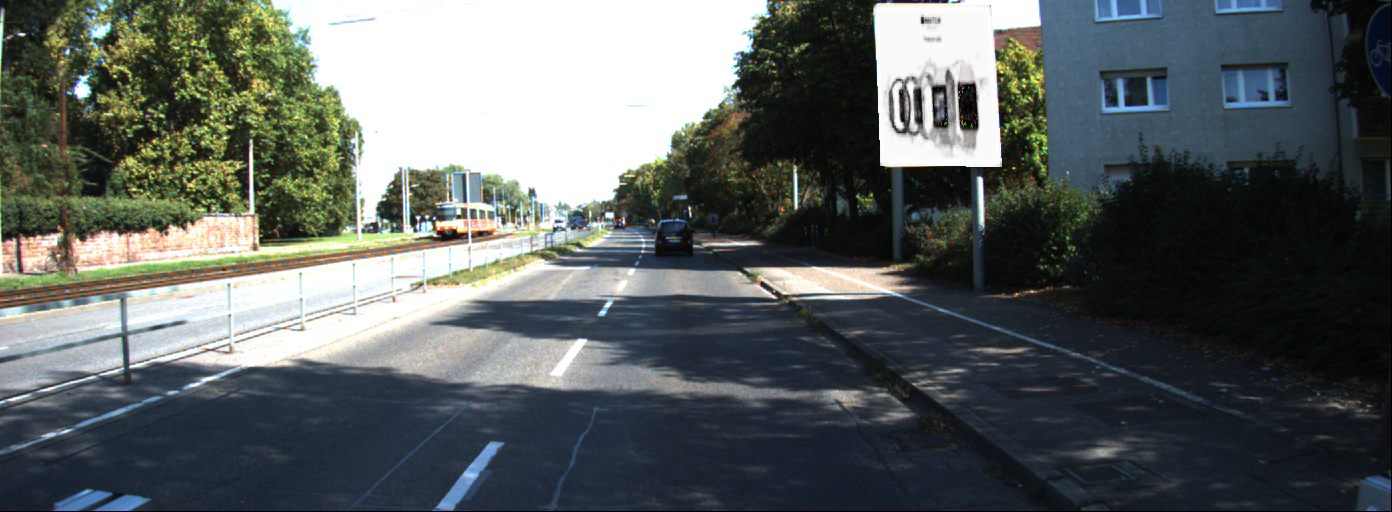} &
        \includefigure{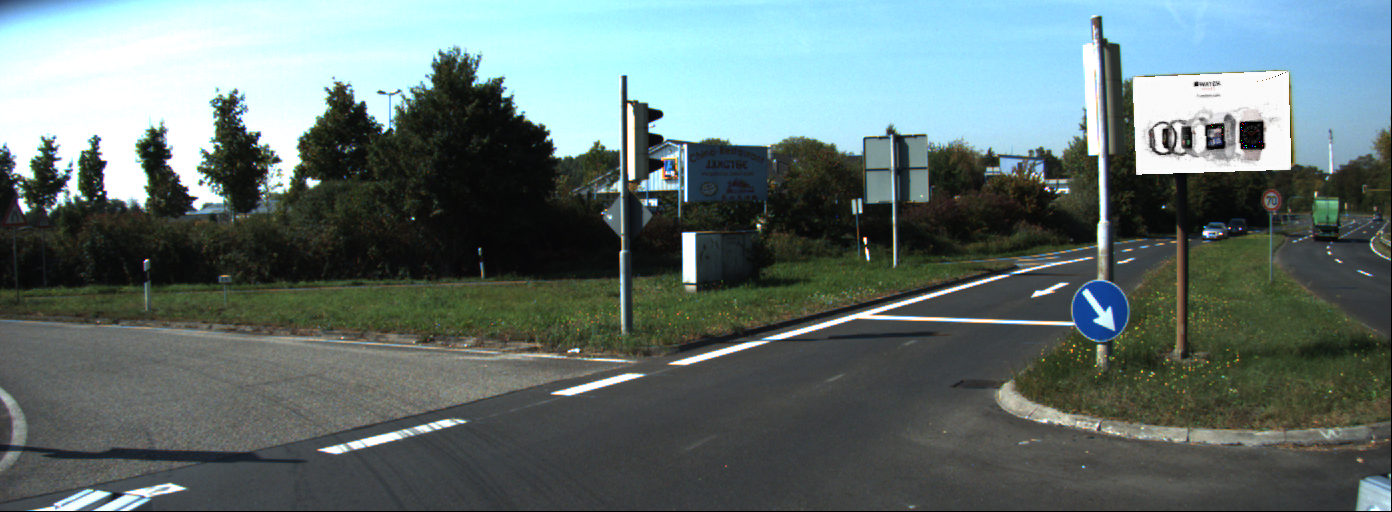} &
        \includefigure{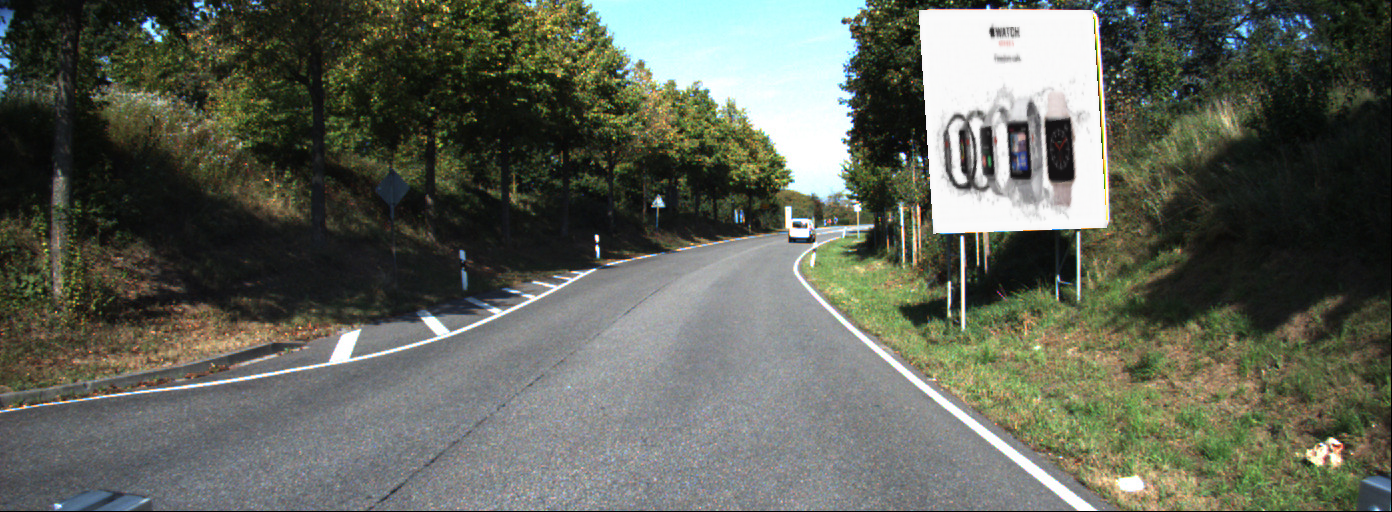} \\
        \hline
        \rotatebox[origin=c]{90}{McDonald's} &
        \includefigure{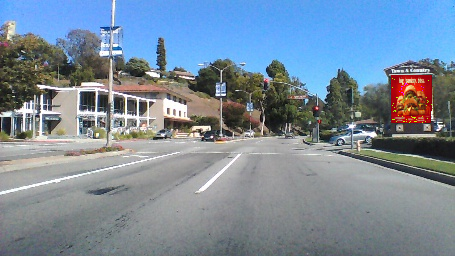} &
        \includefigure{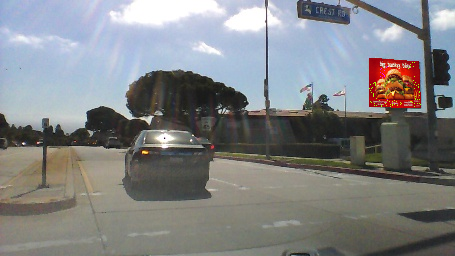} &
        \includefigure{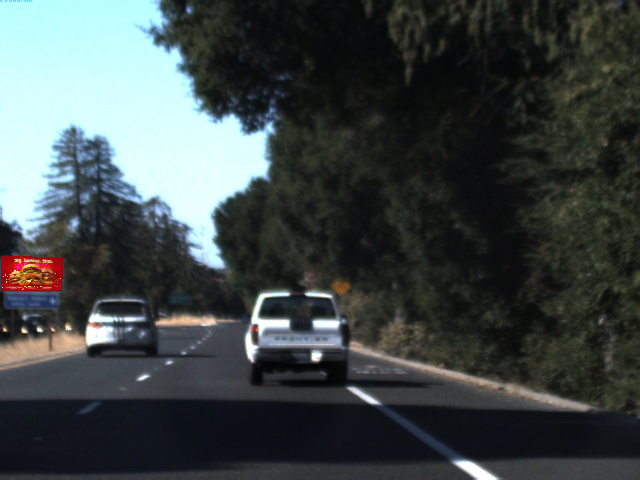} &
        \includefigure{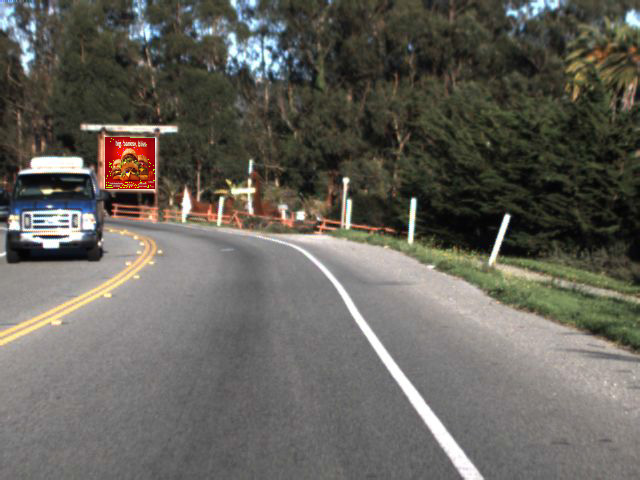} &
        \includefigure{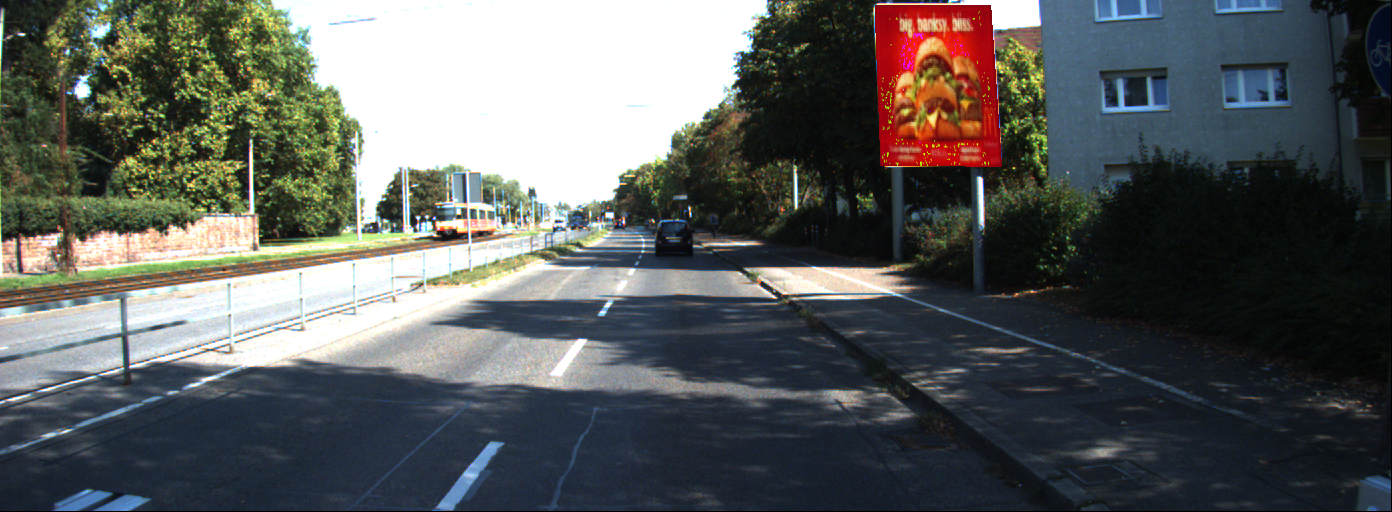} &
        \includefigure{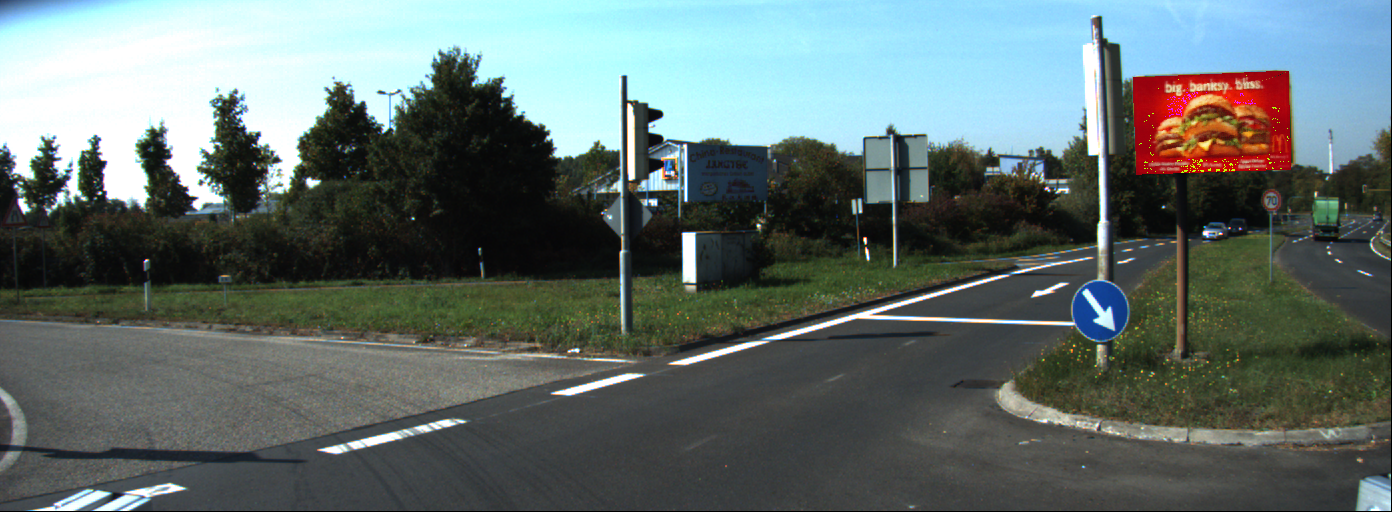} &
        \includefigure{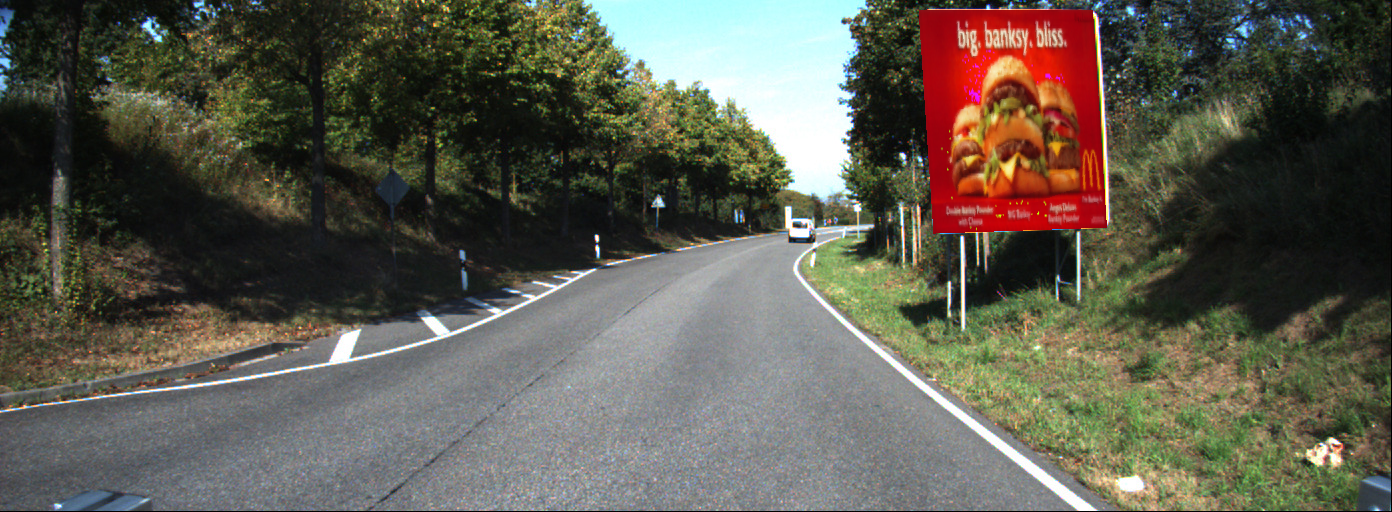} \\
        \hline
    \end{tabular}
    \caption{The original and the generated adversarial fragments and the corresponding image frames under various scenes.}
    \label{tb:overall_digital}
    \vspace{-16pt}
\end{table*}

\begin{figure*}[]
    \centering
    \subfloat[Dave Straight1]{\includegraphics[width=0.245\textwidth]{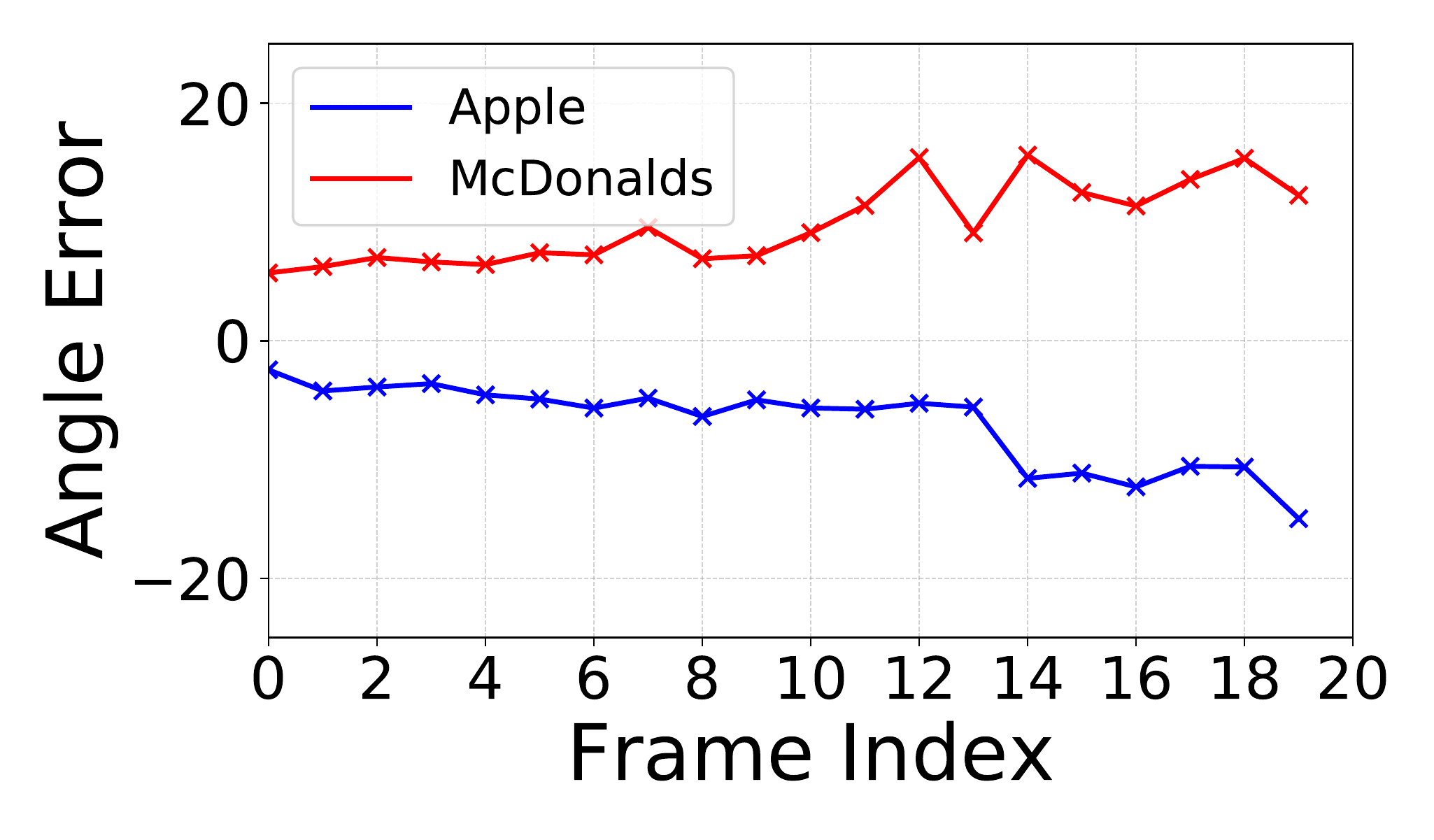}}
    \subfloat[Dave Curve1]{\includegraphics[width=0.245\textwidth]{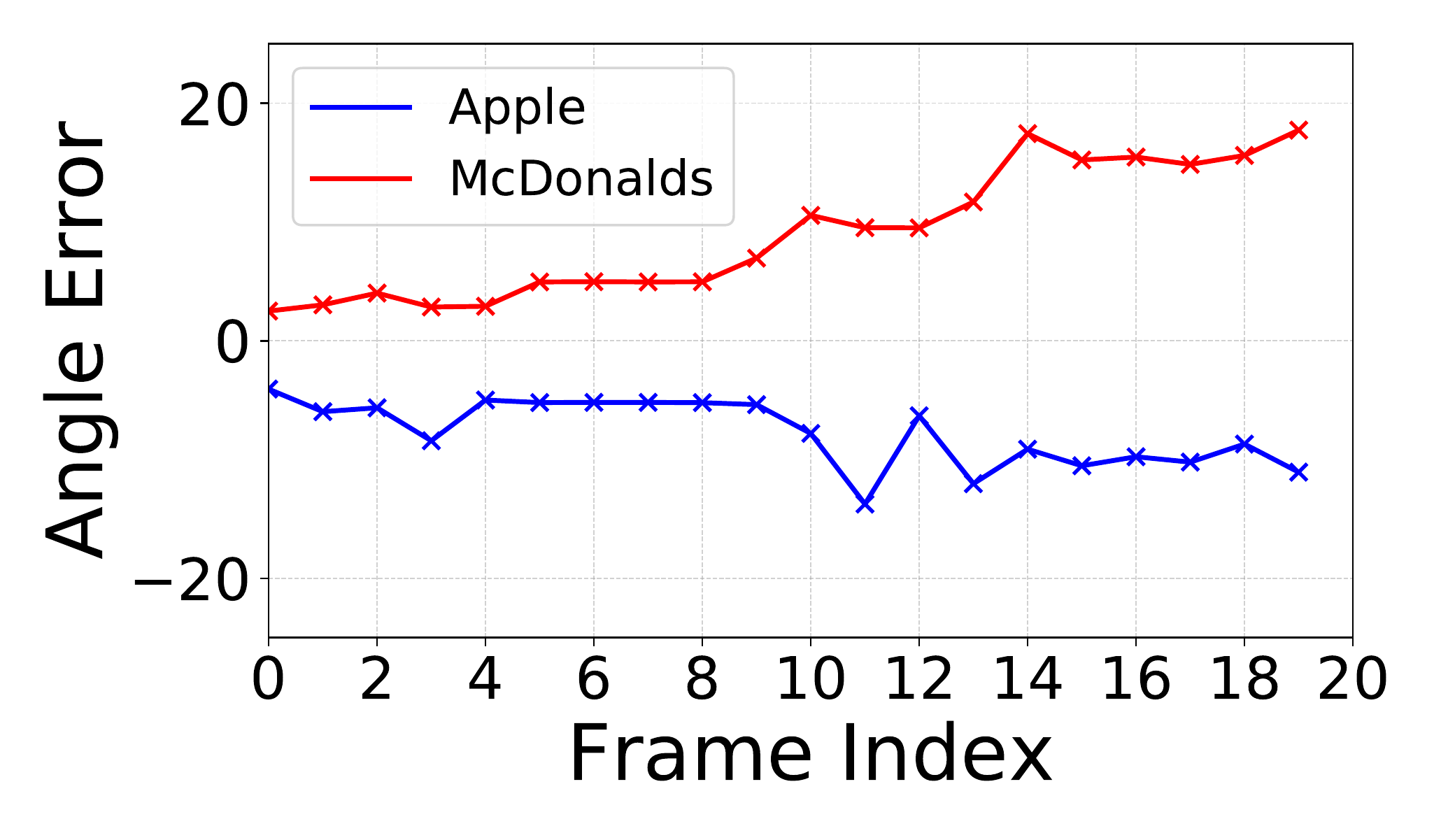}}
    \subfloat[Udacity Straight1]{\includegraphics[width=0.245\textwidth]{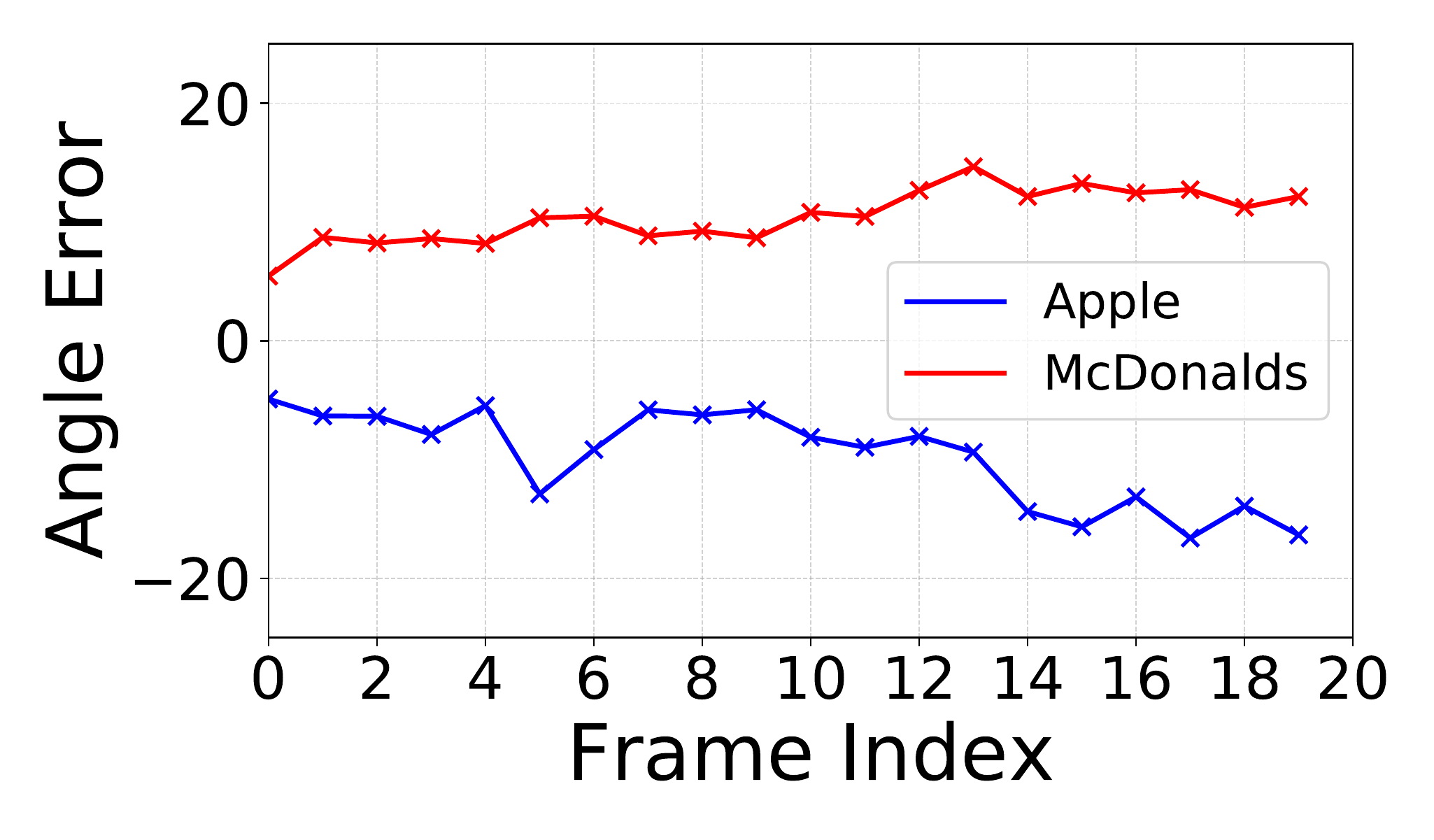}}
    
    \subfloat[Udacity Curve1]{\includegraphics[width=0.245\textwidth]{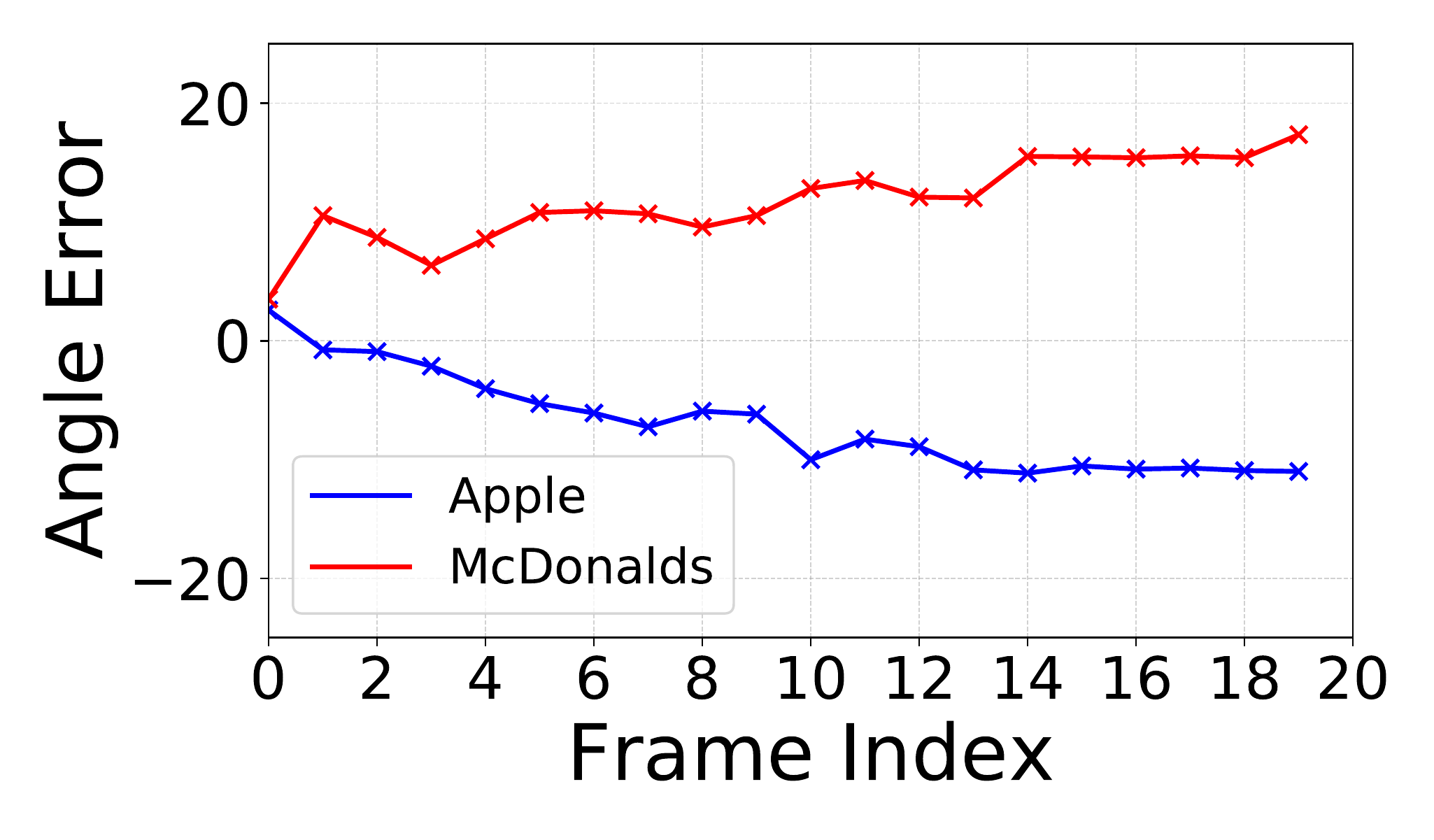}}
    \subfloat[Kitti Straight1]{\includegraphics[width=0.245\textwidth]{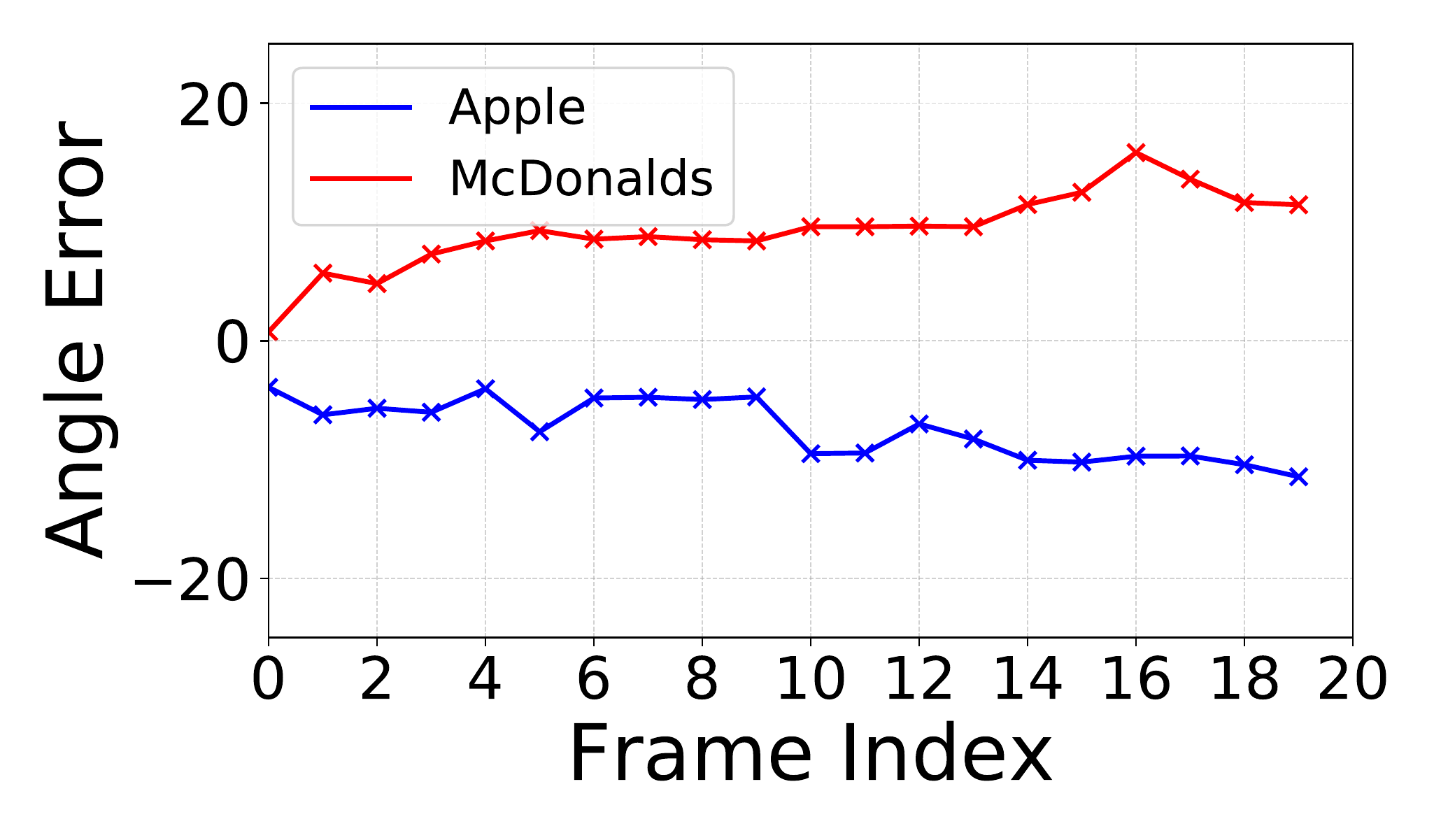}}
    \subfloat[Kitti Straight2]{\includegraphics[width=0.245\textwidth]{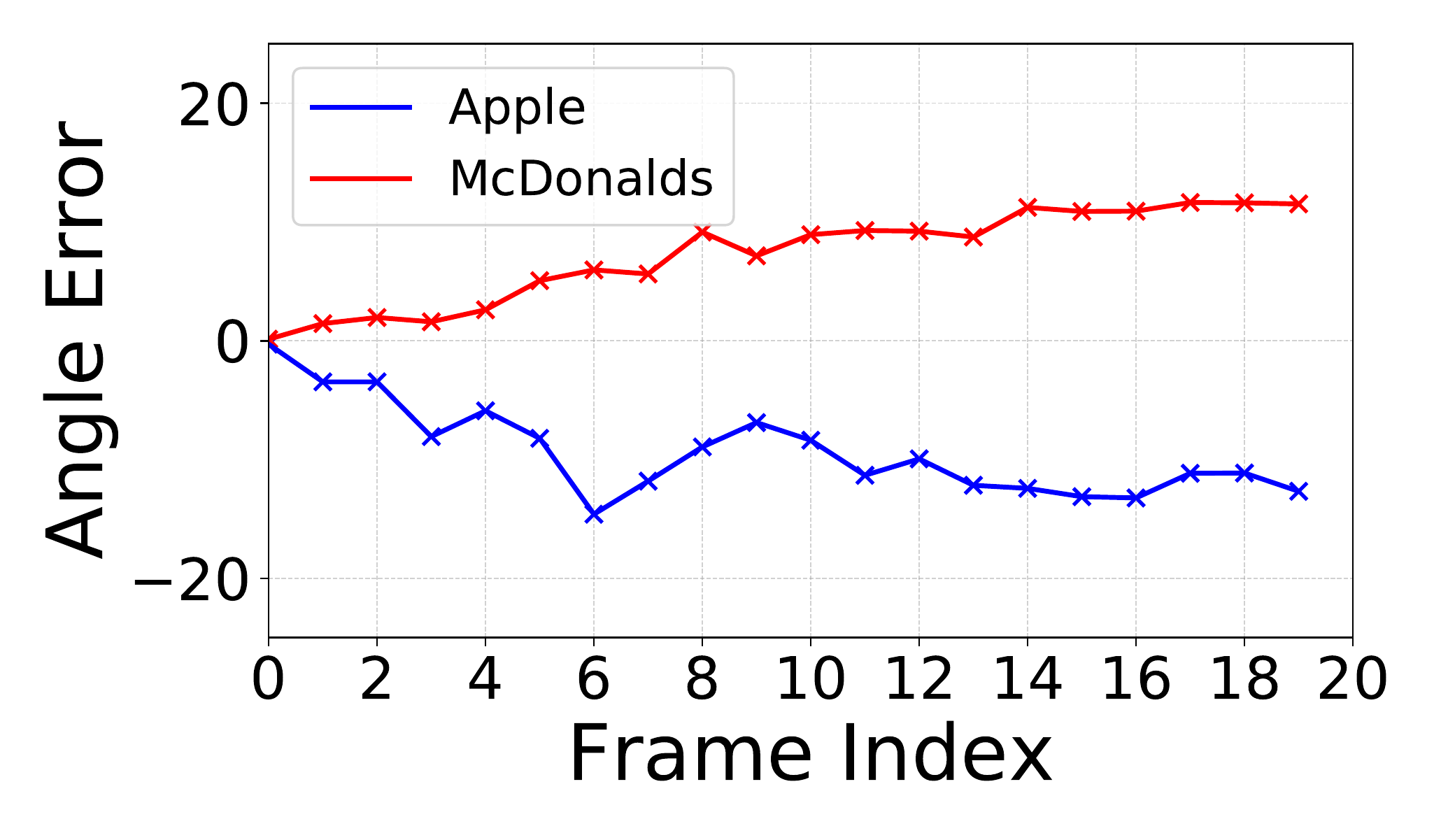}}
    \subfloat[Kitti Curve1]{\includegraphics[width=0.245\textwidth]{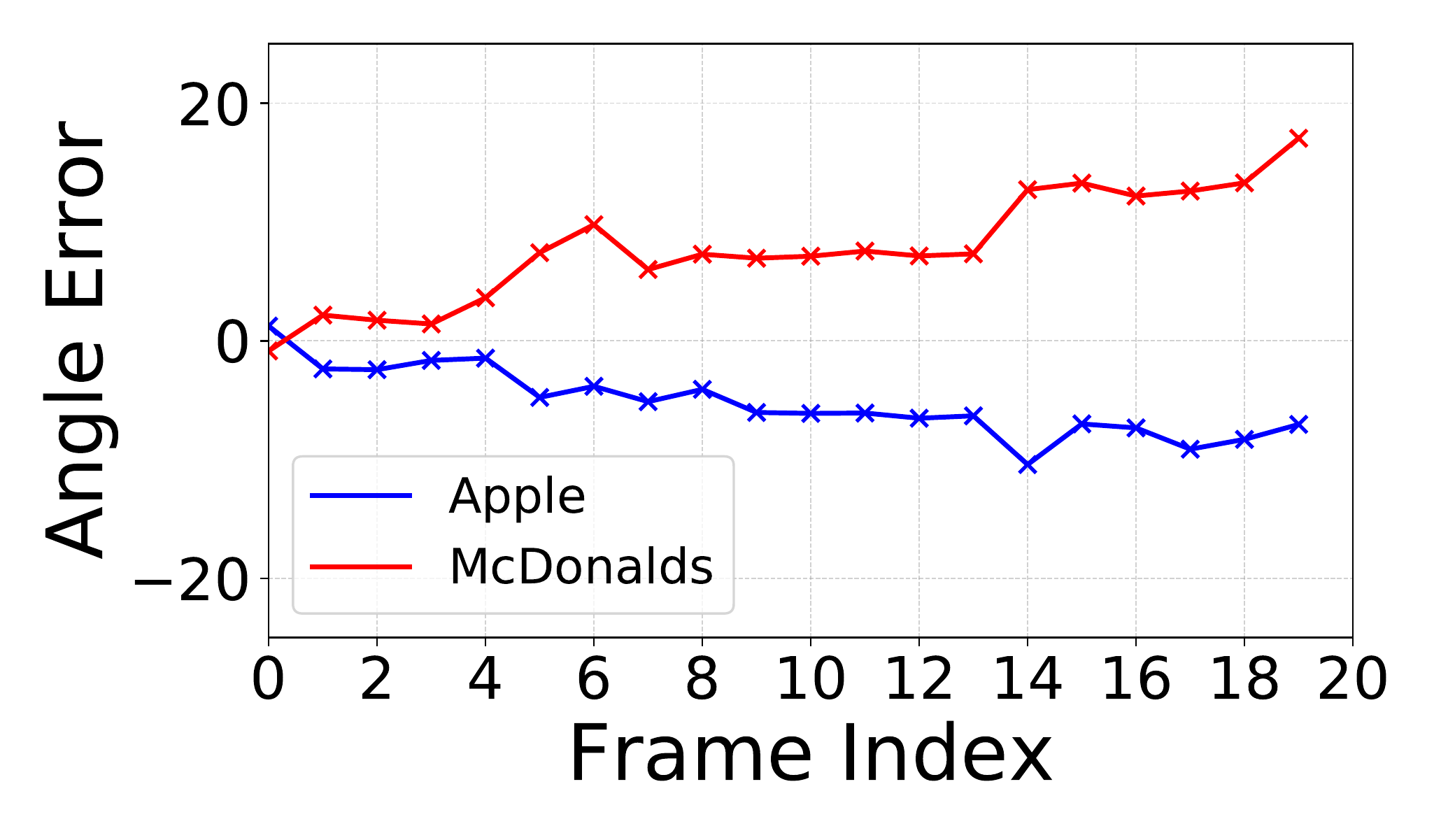}}
    \caption{The illustration of Steering angle error variations along the timeline on steering model Nvidia Dave-2.}
    \label{fig:digital_timeline}
    \vspace{-16pt}
\end{figure*}

\textbf{Baseline methods.}
We compare \name{} with several baseline approaches:

\vspace{-\topsep}
\begin{itemize}
\itemsep 0pt
\item \noindent{\textit{Original sign.}} The first baseline is to simply test the original roadside sign. This comparison is important as it verifies whether steering angle errors are due to \name{} but not the original sign. We include this baseline in both digital and physical evaluations. 

\item \noindent{\textit{FGSM.}}
FGSM~\cite{Goodfellow} is remarkably powerful and it is designed to attack neural networks by leveraging the gradients. In our problem context, we directly apply FGSM to generate perturbations given a captured input frame. We only include FGSM in our digital evaluation, as it is impossible to apply the generated perturbations which covers the entire image frame (\eg, the sky) in physical world.

\item \noindent{\textit{PhysFGSM.}}
In order to apply FGSM in a physical-world setting, we develop a new method called PhysFGSM as an additional baseline, which is based on FGSM and only generate perturbations for the targeted roadside sign in an input image. Doing so allows us to print the perturbed image and paste it onto the corresponding sign. We include PhysFGSM in both digital and physical evaluations. Since the video slice contains multiple frames, PhysFGSM generate perturbations based upon the middle frame.

\item \noindent{\textit{RP2.}} 
We also compare \name{} to a physical-world baseline, RP2~\cite{eykholt2018}, which is an optimization approach that generated perturbations for a single input scene. The original RP2 method focuses on classification problems, so we extend it to be applicable to the steering module by substituting the classification loss with the regression loss.

\item \noindent{\textit{Random Noise.}}
We also print an image perturbed with random noise and paste it on top of the roadside sign.

\end{itemize}

\textbf{Evaluation metrics.}
In our experiments, we use two metrics to evaluate the  efficacy of \name{}: steering angle mean square error (denoted \textit{steering angle MSE}), and \textit{max steering angle error} (MSAE).
\textit{Steering angle MSE} measures the average of the squares of the error between the predicted steering angle and the ground truth, and \textit{MSAE} denotes the maximum steering angle error observed among all frames belonging to a video slice.
A large \textit{steering angle MSE} and \textit{MSAE} implies better attack efficacy.

In addition, we perform online driving testing case studies where we manually control the steering angle in each frame (approximately) according to the real-time calculation of the resulting steering angle error under each evaluated approach. We use the metric \textit{time-to-curb} herein to measure the attack efficacy, which measures the amount of time an actual autonomous driving vehicle would take to drive onto the roadside curb. Please be advised that all the results are relative to the ground truth steering angle.

\renewcommand{\includefigure}[1]{\begin{minipage}{84pt}\vspace{0.5pt}\includegraphics[width=84pt,height=64pt]{#1}\vspace{0.5pt}\end{minipage}}

\begin{table*}[t]
    \centering
    \begin{tabular}{c|c|c||c|c|c}
        \hline
        Original & \name & \name{} frame & \multicolumn{2}{c|}{Baselines} & Baseline frame \\
        \hline
        \includefigure{figures/physical/mc_original.png} &
        \includefigure{figures/physical/mc_generated.png} &
        \includefigure{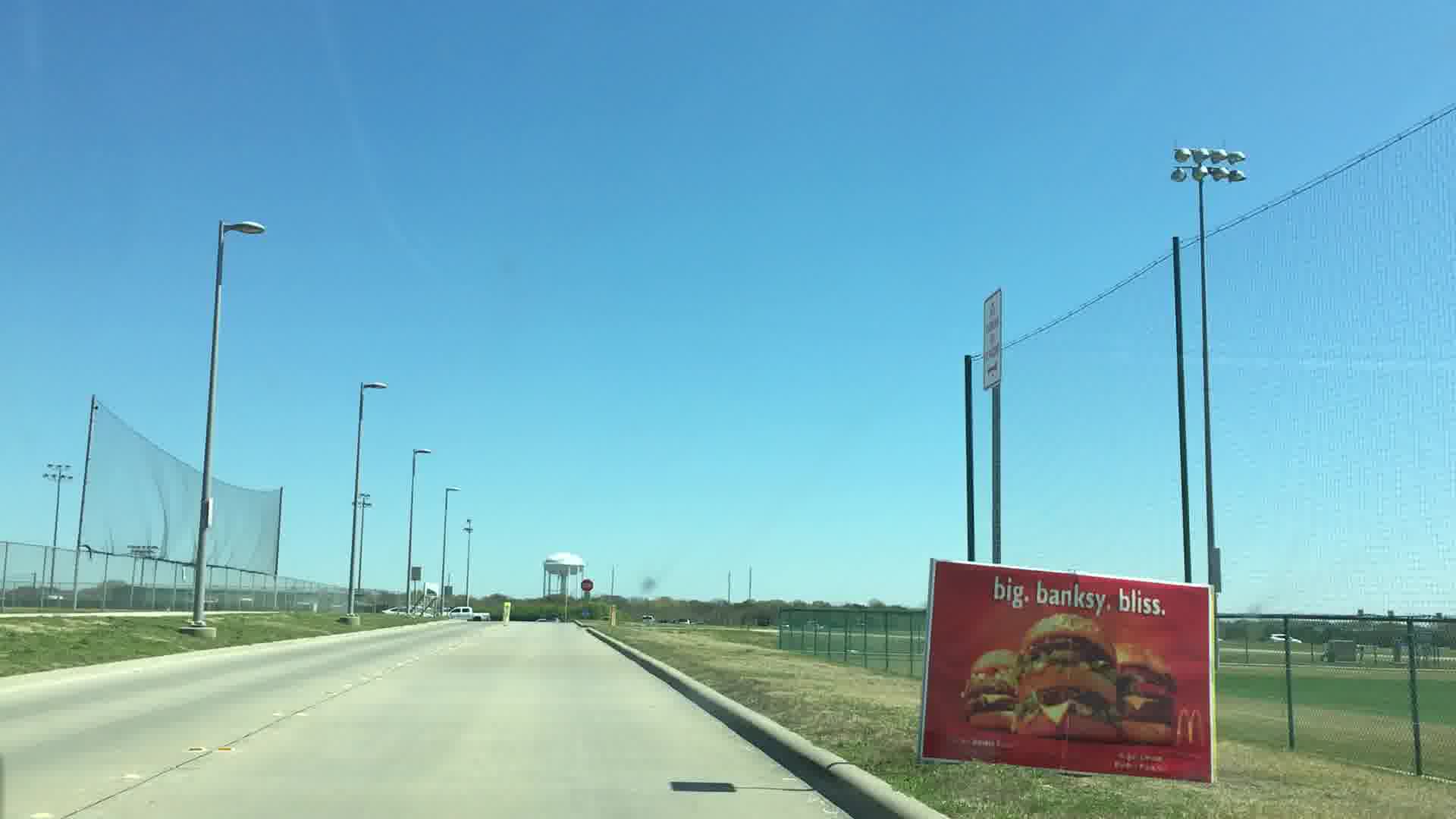} &
        \rotatebox[origin=c]{90}{RP2} &
        \includefigure{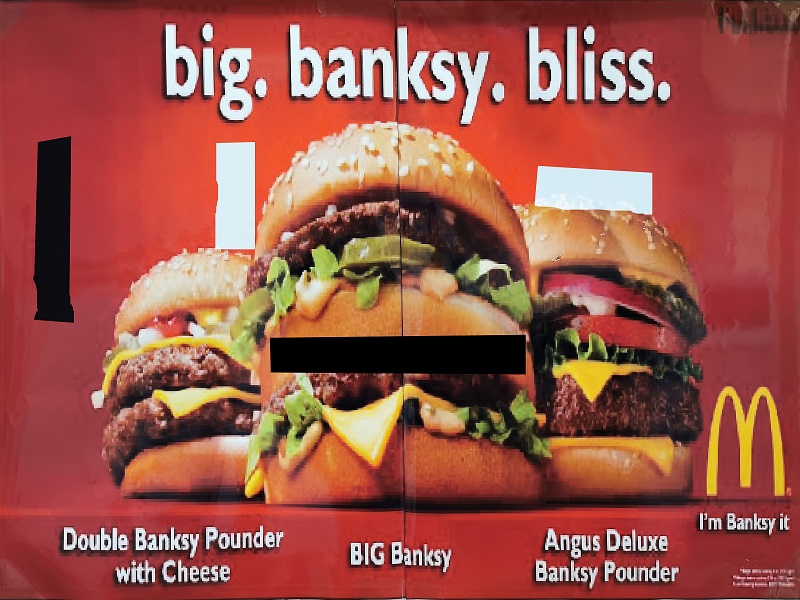} & 
        \includefigure{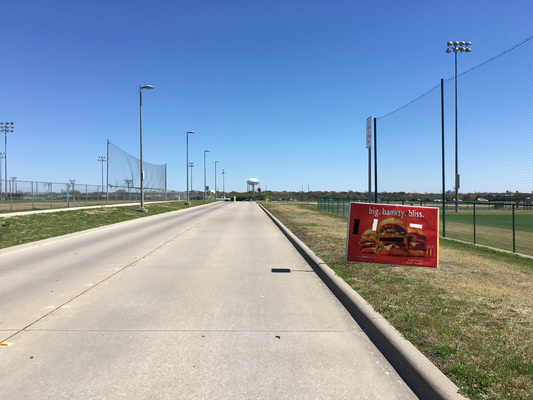}\\
        \hline
        \includefigure{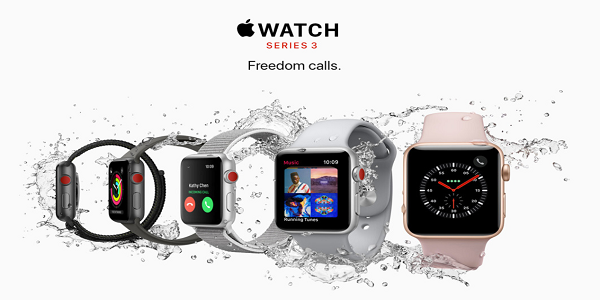} &
        \includefigure{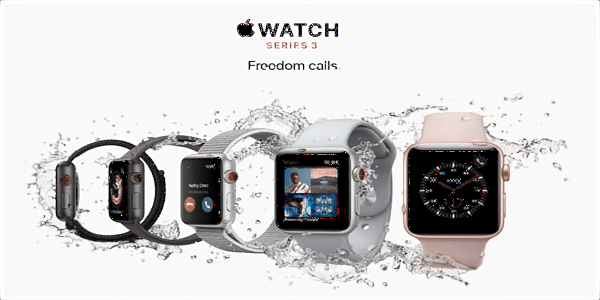} &
        \includefigure{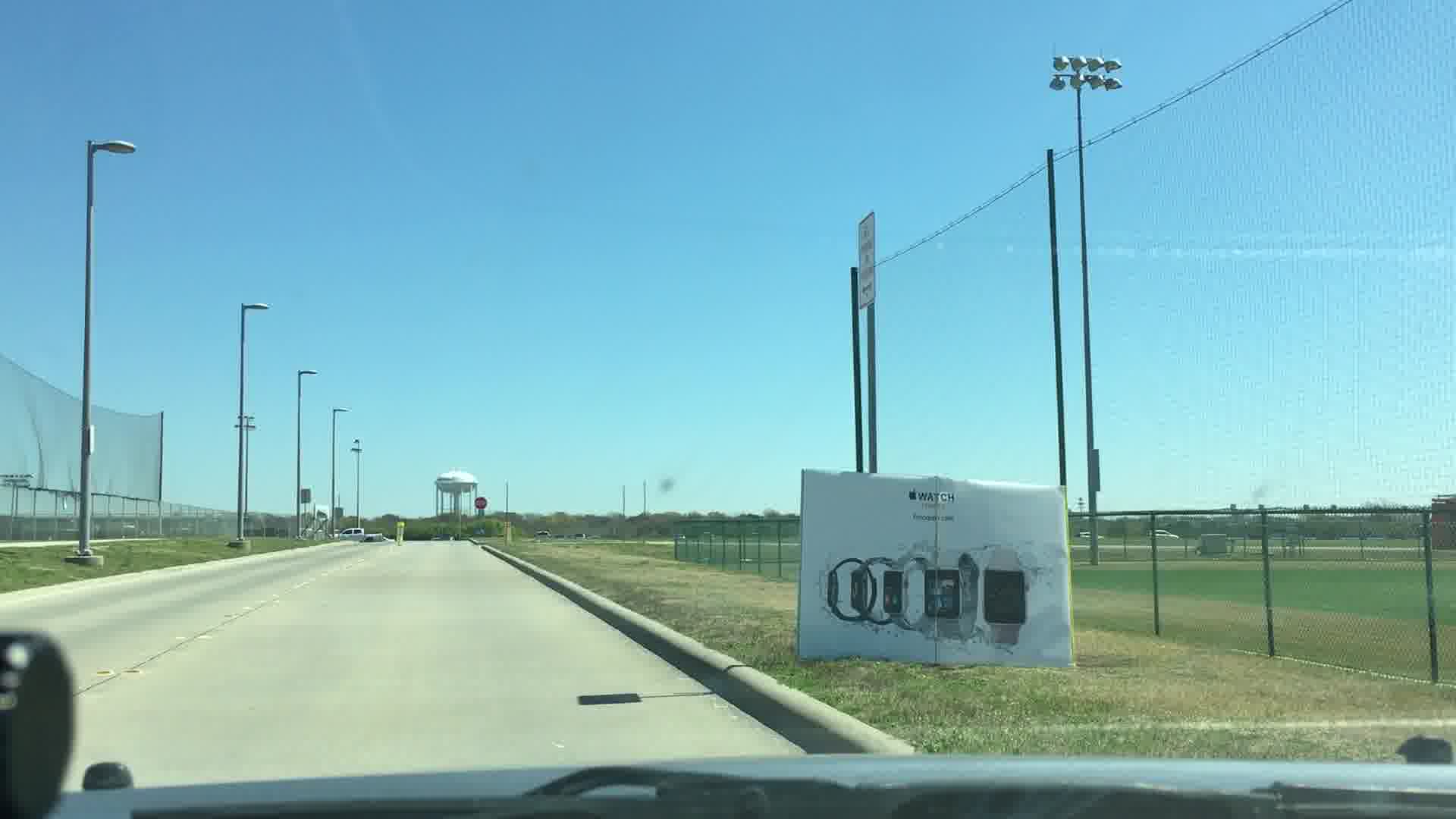} &
        \rotatebox[origin=c]{90}{Random Noise} &
        \includefigure{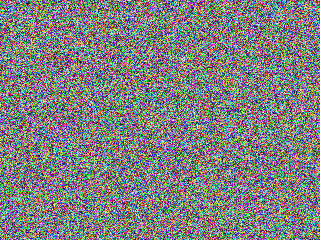} &
        \includefigure{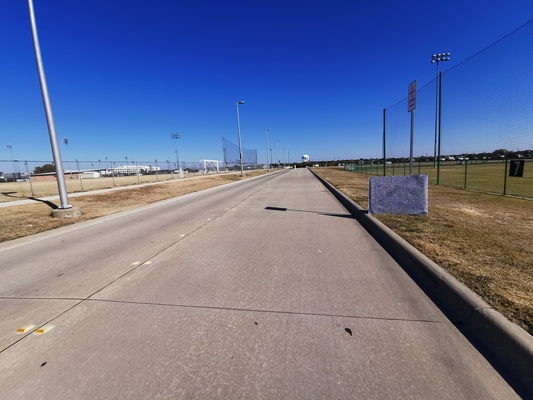} \\
        \hline
    \end{tabular}
    \caption{Illustration of physical-world adversarial scenarios under different approaches.}
    \label{tb:overall_physical}
\end{table*}

\begin{table*}[h]
    \centering
    \resizebox{1.0\textwidth}{!}{
    \begin{tabular}{|l|c|c|c|c|c|c|c|c|c|c|c|c|c|c|c|c|c|c|c|c|}
        \hline
        Frame \#& 1 & 2 & 3 & 4 & 5 & 6 & 7 & 8 & 9 & 10 & 11 & 12 & 13 & 14 & 15 & 16 & 17 & 18 & 19 & 20 \\
        \hline\hline
        Original Apple Sign & 0.36 & -0.51 & 0.82 & 0.45 & 0.10 & -0.16 & 0.84 & -1.38 & -2.16 & -0.86 & 0.60 & -1.11 & 0.21 & -0.49 & -0.55 & -0.56 & 0.10 & -0.51 & 0.49 & -1.00 \\
        \hline
        \name{} (Apple) & 0.17 & 0.89 & 1.68 & 7.94 & 1.93 & 4.79 & 2.87 & 6.34 & 2.08 & 3.54 & 9.06 & 8.37 & 5.93 & 12.51 & 13.43 & 11.37 & 12.75 & 11.74 & 13.63 & 13.44 \\
        \hline\hline
        Original McDonald's Sign & -0.17 & -0.42 & -1.49 & -1.34 & -0.51 & -0.08 & 0.60 & -0.35 & 0.70 & -0.75 & -0.43 & -0.35 & 0.59 & -0.89 & 1.49 & 0.61 & 0.94 & -0.99 & 1.13 & -0.00 \\
        \hline
        \name{} (McDonald's) & -1.24 & -1.37 & -0.02 & -0.30 & -2.48 & -0.17 & -1.06 & -0.80 & -0.01 & -5.37 & -1.60 & -2.62 & -2.45 & -4.68 & -11.71 & -10.85 & -9.83 & -8.74 & -11.35 & -19.17 \\
        \hline
    \end{tabular}
    }
    \caption{Per-frame steering angle error under physical-world experiments. Rows 2 and 4 (rows 3 and 5) show the steering angle error when the original signs (corresponding adversarial signs generated by \name) are deployed.}
    \vspace{-8pt}
    \label{tab:physical_normal_angle_error}
\end{table*}

\subsection{Results}

\begin{table*}[h]
    \centering
    \resizebox{1.0\textwidth}{!}{
        \begin{tabular}{|l|l|c|c|c|c|c|c|c|}
        \hline
        \multirow{2}{*}{Steering Model} & \multirow{2}{*}{Approach} & \multicolumn{2}{c|}{Dave} & \multicolumn{2}{c|}{Udacity} & \multicolumn{3}{c|}{Kitti} \\
        \cline{3-9}
        & & Straight1 & Curve1 & Straight1 & Curve1 & Straight1 & Straight2 & Curve1 \\
        \hline\hline
        \multirow{5}{*}{Nvidia Dave-2} & \name{} & 106.69 / 15.61 & 66.79 / 11.63 & 81.76 / 17.04 & 114.13 / 14.64 & 108.76 / 17.72 & 150.00 / 17.34 & 95.87 / 15.83 \\
        \cline{2-9}
        & FGSM & 115.91 / 17.41 & 199.27 / 19.47 & 141.23 / 16.17 & 192.19 / 21.23 & 156.16 / 17.84 & 217.52 / 19.50 & 103.38 / 14.54 \\
        \cline{2-9}
        & PhysFGSM & 15.88 / 6.42 & 4.73 / 4.87 & 13.91 / 5.74 & 3.08 / 2.89 & 15.17 / 8.04 & 8.67 / 4.54 & 13.12 / 7.24 \\
        \cline{2-9}
        & Random Noise & 3.00 / 2.01 & 2.25 / 2.37 & 2.36 / 2.60 & 1.77 / 3.10 & 3.15 / 3.16 & 1.60 / 0.96 & 5.92 / 4.41 \\
        \cline{2-9}
        & Original Sign & 4.17 / 3.15 & 4.35 / 2.40 & 3.84 / 1.79 & 1.09 / 0.72 & 4.20 / 2.98 & 3.06 / 1.23 & 2.86 / 1.30 \\
        \hline\hline
        \multirow{5}{*}{Udacity Cg23} & \name{} & 91.85 / 13.80 & 113.41 / 14.78 & 50.61 / 10.43 & 78.56 / 15.46 & 46.53 / 11.72 & 62.64 / 11.64 & 71.09 / 18.14 \\
        \cline{2-9}
        & FGSM & 203.34 / 19.70 & 157.98 / 14.67 & 171.92 / 19.89 & 96.74 / 17.75 & 136.08 / 14.00 & 162.35 / 18.53 & 89.75 / 16.71 \\
        \cline{2-9}
        & PhysFGSM & 58.53 / 11.86 & 36.44 / 10.68 & 30.72 / 9.41 & 46.74 / 8.88 & 28.89 / 11.37 & 22.63 / 7.61 & 61.23 / 10.95 \\
        \cline{2-9}
        & Random Noise & 5.32 / 3.67 & 3.75 / 2.72 & 4.05 / 2.52 & 4.20 / 2.26 & 5.31 / 4.49 & 6.54 / 1.98 & 6.10 / 3.68 \\
        \cline{2-9}
        & Original Sign & 4.17 / 3.15 & 4.35 / 2.40 & 3.84 / 1.79 & 4.09 / 2.72 & 4.20 / 2.98 & 3.06 / 1.23 & 2.30 / 1.86 \\
        \hline\hline
        \multirow{5}{*}{Udacity Rambo} & \name{} & 61.87 / 11.28 & 113.78 / 15.29 & 87.68 / 13.90 & 42.71 / 12.55 & 56.41 / 12.42 & 58.67 / 10.42 & 145.66 / 21.85 \\
        \cline{2-9}
        & FGSM & 209.81 / 21.78 & 147.28 / 16.43 & 151.14 / 15.28 & 166.50 / 16.27 & 169.17 / 18.57 & 126.14 / 14.19 & 175.28 / 19.36 \\
        \cline{2-9}
        & PhysFGSM & 16.43 / 8.95 & 14.24 / 8.34 & 5.32 / 3.73 & 14.82 / 6.11 & 16.58 / 7.78 & 13.89 / 7.93 & 29.58 / 19.18 \\
        \cline{2-9}
        & Random Noise & 1.90 / 2.55 & 3.49 / 5.79 & 6.06 / 5.00 & 1.92 / 3.98 & 3.82 / 5.42 & 2.09 / 3.05 & 1.52 / 1.91 \\
        \cline{2-9}
        & Original Sign & 3.93 / 2.01 & 6.30 / 4.46 & 1.80 / 1.28 & 6.54 / 2.52 & 5.06 / 3.52 & 5.75 / 4.03 & 4.95 / 2.07 \\
        \hline
    \end{tabular}
    }
    \caption{\textit{Steering angle MSE} (left) and \textit{MSAE} (right) under all evaluated approaches. Although FGSM produces the maximal attacks, it modifies the whole image observation and is not applicable to the real world. Among all physical-world attack approaches, our approach \name{} produces the best performance.}
    \label{tab:digital_baselines_mse}
    \vspace{-8pt}
\end{table*}

\begin{table*}[h]
    \centering
    \begin{tabular}{|l|c|c|c|c|}
        \hline
        & \name & RP2 & Random Noise & Original Sign \\
        \hline
        Nvidia Dave-2 & 73.94 / 13.63 & 23.48 / 6.52 & 2.48 / 1.02 & 2.12 / 1.56 \\
        \hline
        Udacity Cg23 & 99.23 / 14.56 & 25.15 / 7.86 & 2.56 / 2.11 & 2.15 / 1.73 \\
        \hline
        Udacity Rambo & 87.56 / 17.60 & 32.54 / 7.51 & 1.51 / 1.15 & 3.12 / 2.48 \\
        \hline
    \end{tabular}
    \caption{\textit{Steering angle MSE} (left) and \textit{MSAE} (right) under \name{}, RP2, random noise, and original sign.}
    \label{tab:physical_baselines_mse}
    \vspace{-8pt}
\end{table*}



We first report the overall efficacy under \name{} in both digital and physical-world scenarios. A complete set of results is given in the supplementary document.

\textbf{Results on digital scenarios.} 
Table~\ref{tb:overall_digital} shows a representative frame of each scene where the signs are replaced with adversarial examples generated from \name{} (using the targeted steering model NVIDIA Dave-2).
Each column of Table~\ref{tb:overall_digital} represents a specific scene. It is observed that \name{} can generate rather realistic adversarial samples, visually indistinguishable from the original objects. 
The targeted roadside signs in the original video slices are replaced by our selected McDonald and Apple Watch signs, and the modified video slices are used in all experiments. This is because the roadside signs in the original video slices have a low resolution, which makes it hard to verify whether our generated \advobjects{} are visually distinguishable.

Fig.~\ref{fig:digital_timeline} shows the results on steering angle error along the timeline in each frame scene, where the size of the adversarial image increases nearly monotonically over time. 
Each sub-figure in Fig.~\ref{fig:digital_timeline} indicates a specific scene, where the x-axis represents the frame index along the timeline, and the y-axis represents the steering angle error. We clearly observe that \name{} leads to noticeable angle error for almost all frames, even for earlier frames in which the adversarial sample is relatively small compared to the background. 

\textbf{Results on physical-world scenarios.}
We perform physical-world experiments as follows. We first record training videos of driving a vehicle towards the original roadside sign and use these videos to train the Dave-2 model. We then train \name{} following the same configuration as the digital-world evaluation to generate adversarial samples. The generated adversarial samples was then printed and pasted on the original roadside sign. We then recorded testing videos of the same drive-by process but with the adversarial sample. The steering angle error are then obtained by analyzing these testing videos. 
Specifically, for both training and testing video slices, we start recording at 70 ft away and stop recording when the vehicle physically passes the advertising sign. For training videos, the driving speed is set to be $10 mph$ to capture sufficient images. The speed for the testing video is set to be $20 mph$ to reflect ordinary on-campus driving speed limit. The physical case studies are performed on a straight lane due to safety reasons. The size of the roadside advertising board used in our experiment is $48' \times 72'$.


Table~\ref{tb:overall_physical} shows the original sign and the corresponding adversarial examples generated under different methods as well as a camera-captured scenes for each example. 
To clearly interpret the results, we list the per-frame steering angle error due to \name{} and using the original sign in Table~\ref{tab:physical_normal_angle_error} (additional comparison results are detailed in Sec.~\ref{sec:baseline}).
As seen in Table~\ref{tab:physical_normal_angle_error},  \name{} is able to generate a single printable physical-world-resilient adversarial example which could mislead the driving model for continuous frames during the entire driving process. An interesting observation herein is that the steering angle error tends to increase along with the increased frame index. This is because, with a larger frame index, the size of the adversarial sample occupies a relatively large space in the entire frame, so being able to more negatively impact the steering model. Also, we observe that with the original roadside sign, the steering angle error is almost negligible under all frames.

\subsection{Comparison against Baseline Approaches}
\label{sec:baseline}

\textbf{Digital Baselines.}
For each steering model, we compare our approach with four other baselines including FGSM, PhysFGSM, random noise, and original sign. 
Table~\ref{tab:digital_baselines_mse} shows the results  on seven different scenes.
These results suggest the following observations: 
(1) although FGSM achieves the highest attacking effect, it needs to apply perturbations to the entire scene, which is not applicable to the physical world;
(2) the attacking effectiveness of our approach is much better than PhysFGSM, implying that once considering physical world implementation constraints, \name{} would outperform direct extensions of existing approaches. 
(3) each steering model is reasonably robust as the angle errors under random noise and original sign are trivial.

\textbf{Physical Baselines.}
For physical-world scenarios, we compare \name{} against PhysFGSM, random noise, and original sign. The results are shown in Table~\ref{tab:physical_baselines_mse}. 
We observe that both random noise and original sign have negligible impact on the steering models, which indicate the pre-trained steering models (without being attacked) are  sufficiently robust in physical world settings. 
As seen in Table~\ref{tab:physical_baselines_mse}, \name{} significantly outperforms RP2 and can achieve very high attack efficacy under all steering models, which may lead to dangerous driving actions in the real world.


\begin{table}[h]
    \resizebox{1.0\columnwidth}{!}{
    \begin{tabular}{|l|c|c|c|c|}
        \hline
        & \name & RP2 & Random Noise & Original \\
        \hline
        Time-to-curb & $10s$ & - & - & - \\
        \hline
        Distance-to-center & 1.5m & 1.09m & 0.29m & 0.47m \\
        \hline
    \end{tabular}
    }
    \caption{Online driving testing results. The second row shows the time-to-curb result and the third row shows the maximum distance that the vehicle deviates from the correct path (\ie, driving straight).}
    \label{tab:online_test_diagram}
    \vspace{-8pt}
\end{table}

\subsection{Online Driving Case Studies}

The above evaluations are off-policy where the driving trajectory was not affected by the adversarial signs. In this section, we further conduct on-policy evaluation, \ie, online driving case studies mimicking the actual driving scenario to learn how would \name{} impact the steering decision by an actual autonomous vehicle. In these case studies, we manually control steering in a real-time fashion within each frame according to the calculated steering angle error under each approach with the steering model Nvidia Dave-2. We ask a human driver to drive the vehicle at $5 mph$ for 1 second to reflect one frame and a corresponding manual steering action. 
We note that this online evaluation setup is a proxy of real autonomous vehicle and provides proper evaluation of an attack system. We do not use virtual simulators for evaluation because they normally causes sim-to-real transfer issues. So the evaluation results on a simulator would not reflect the model’s capability in the physical world.
As seen in Table~\ref{tab:online_test_diagram}, \name{} outperforms the other baselines under the two metrics. Also, only the adversarial sign generated under \name{} leads the vehicle to drive onto the roadside curb, which takes 10s (given the very low driving speed due to safety concerns). This online driving case study further demonstrates the dangerous steering action an autonomous vehicle would take due to \name{}, indicating its effectiveness when applied to actual autonomous vehicles.





\section{Conclusion}
We present \name{}, which generates physical-world-resilient adversarial examples for misleading autonomous steering systems. 
We proposed a novel GAN-based framework for generating a single adversarial example that continuously misleads the driving model during the entire trajectory. The generated adversarial example is visually indistinguishable from the original roadside object. Extensive digital and physical-world experiments show the efficacy and robustness of \name{}. We hope our work could inspire future research on safe and robust machine learning for autonomous driving. 

{\small
\bibliographystyle{ieee_fullname}
\bibliography{main}

\begin{thebibliography}{10}\itemsep=-1pt

\bibitem{persepective_mapping}
Perspectives mapping,
  \url{https://www.geometrictools.com/Documentation/PerspectiveMappings.pdf}.

\bibitem{athalye2017}
Anish Athalye, Logan Engstrom, Andrew Ilyas, and Kevin Kwok.
\newblock Synthesizing robust adversarial examples.
\newblock {\em arXiv preprint arXiv:1707.07397}, 2017.

\bibitem{broggi2013}
Alberto Broggi, Michele Buzzoni, Stefano Debattisti, Paolo Grisleri,
  Maria~Chiara Laghi, Paolo Medici, and Pietro Versari.
\newblock Extensive tests of autonomous driving technologies.
\newblock {\em IEEE Transactions on Intelligent Transportation Systems},
  14(3):1403--1415, 2013.

\bibitem{carlini2017}
Nicholas Carlini and David Wagner.
\newblock Towards evaluating the robustness of neural networks.
\newblock In {\em 2017 IEEE Symposium on Security and Privacy (SP)}, pages
  39--57. IEEE, 2017.

\bibitem{carlini2017b}
Nicholas Carlini and David Wagner.
\newblock Towards evaluating the robustness of neural networks.
\newblock In {\em 2017 IEEE Symposium on Security and Privacy (SP)}, pages
  39--57. IEEE, 2017.

\bibitem{chen2015}
Chenyi Chen, Ari Seff, Alain Kornhauser, and Jianxiong Xiao.
\newblock Deepdriving: Learning affordance for direct perception in autonomous
  driving.
\newblock In {\em Proceedings of the IEEE International Conference on Computer
  Vision}, pages 2722--2730, 2015.

\bibitem{elsayed2018}
Gamaleldin Elsayed, Shreya Shankar, Brian Cheung, Nicolas Papernot, Alexey
  Kurakin, Ian Goodfellow, and Jascha Sohl-Dickstein.
\newblock Adversarial examples that fool both computer vision and time-limited
  humans.
\newblock In {\em Advances in Neural Information Processing Systems}, pages
  3914--3924, 2018.

\bibitem{eykholt2018}
Kevin Eykholt, Ivan Evtimov, Earlence Fernandes, Bo Li, Amir Rahmati, Chaowei
  Xiao, Atul Prakash, Tadayoshi Kohno, and Dawn Song.
\newblock Robust physical-world attacks on deep learning visual classification.
\newblock In {\em CVPR}, pages 1625--1634, 2018.

\bibitem{geiger2013vision}
Andreas Geiger, Philip Lenz, Christoph Stiller, and Raquel Urtasun.
\newblock Vision meets robotics: The kitti dataset.
\newblock {\em The International Journal of Robotics Research},
  32(11):1231--1237, 2013.

\bibitem{GAN}
Ian Goodfellow, Jean Pouget-Abadie, Mehdi Mirza, Bing Xu, David Warde-Farley,
  Sherjil Ozair, Aaron Courville, and Yoshua Bengio.
\newblock Generative adversarial nets.
\newblock In {\em NIPS}, pages 2672--2680. 2014.

\bibitem{Goodfellow}
Ian Goodfellow, Jonathon Shlens, and Christian Szegedy.
\newblock Explaining and harnessing adversarial examples.
\newblock In {\em ICLR}, 2015.

\bibitem{kurakin2016}
Alexey Kurakin, Ian Goodfellow, and Samy Bengio.
\newblock Adversarial examples in the physical world.
\newblock {\em arXiv preprint arXiv:1607.02533}, 2016.

\bibitem{kurakin2017}
Alexey Kurakin, Ian Goodfellow, and Samy Bengio.
\newblock Adversarial machine learning at scale.
\newblock {\em ICLR}, 2017.

\bibitem{liu2016delving}
Yanpei Liu, Xinyun Chen, Chang Liu, and Dawn Song.
\newblock Delving into transferable adversarial examples and black-box attacks.
\newblock {\em ICLR}, 2017.

\bibitem{lu2017no}
Jiajun Lu, Hussein Sibai, Evan Fabry, and David Forsyth.
\newblock No need to worry about adversarial examples in object detection in
  autonomous vehicles.
\newblock {\em arXiv preprint arXiv:1707.03501}, 2017.

\bibitem{lu2017conditional}
Yongyi Lu, Yu-Wing Tai, and Chi-Keung Tang.
\newblock Conditional cyclegan for attribute guided face image generation.
\newblock {\em arXiv preprint arXiv:1705.09966}, 2017.

\bibitem{metzen2017}
Jan~Hendrik Metzen, Mummadi~Chaithanya Kumar, Thomas Brox, and Volker Fischer.
\newblock Universal adversarial perturbations against semantic image
  segmentation.
\newblock In {\em 2017 IEEE International Conference on Computer Vision
  (ICCV)}, pages 2774--2783. IEEE, 2017.

\bibitem{mogelmose2012}
Andreas Mogelmose, Mohan~Manubhai Trivedi, and Thomas~B Moeslund.
\newblock Vision-based traffic sign detection and analysis for intelligent
  driver assistance systems: Perspectives and survey.
\newblock {\em IEEE Transactions on Intelligent Transportation Systems},
  13(4):1484--1497, 2012.

\bibitem{moosavi2016}
Seyed-Mohsen Moosavi-Dezfooli, Alhussein Fawzi, and Pascal Frossard.
\newblock Deepfool: a simple and accurate method to fool deep neural networks.
\newblock In {\em CVPR}, pages 2574--2582, 2016.

\bibitem{pan2017virtual}
Xinlei Pan, Yurong You, Ziyan Wang, and Cewu Lu.
\newblock Virtual to real reinforcement learning for autonomous driving.
\newblock {\em arXiv preprint arXiv:1704.03952}, 2017.

\bibitem{papernot2017}
Nicolas Papernot, Patrick McDaniel, Ian Goodfellow, Somesh Jha, Z~Berkay Celik,
  and Ananthram Swami.
\newblock Practical black-box attacks against machine learning.
\newblock In {\em Proceedings of the 2017 ACM on Asia Conference on Computer
  and Communications Security}, pages 506--519. ACM, 2017.

\bibitem{papernot2016}
Nicolas Papernot, Patrick McDaniel, Somesh Jha, Matt Fredrikson, Z~Berkay
  Celik, and Ananthram Swami.
\newblock The limitations of deep learning in adversarial settings.
\newblock In {\em 2016 IEEE European Symposium on Security and Privacy
  (EuroS\&P)}, pages 372--387. IEEE, 2016.

\bibitem{sermanet2011}
Pierre Sermanet and Yann LeCun.
\newblock Traffic sign recognition with multi-scale convolutional networks.
\newblock In {\em IJCNN}, pages 2809--2813, 2011.

\bibitem{sharif2016}
Mahmood Sharif, Sruti Bhagavatula, Lujo Bauer, and Michael~K Reiter.
\newblock Accessorize to a crime: Real and stealthy attacks on state-of-the-art
  face recognition.
\newblock In {\em Proceedings of the 2016 ACM SIGSAC Conference on Computer and
  Communications Security}, pages 1528--1540. ACM, 2016.

\bibitem{szegedy2013}
Christian Szegedy, Wojciech Zaremba, Ilya Sutskever, Joan Bruna, Dumitru Erhan,
  Ian Goodfellow, and Rob Fergus.
\newblock Intriguing properties of neural networks.
\newblock {\em arXiv preprint arXiv:1312.6199}, 2013.

\bibitem{tanay2016boundary}
Thomas Tanay and Lewis Griffin.
\newblock A boundary tilting persepective on the phenomenon of adversarial
  examples.
\newblock {\em arXiv preprint arXiv:1608.07690}, 2016.

\bibitem{tian2018}
Yuchi Tian, Kexin Pei, Suman Jana, and Baishakhi Ray.
\newblock Deeptest: Automated testing of deep-neural-network-driven autonomous
  cars.
\newblock In {\em ICSE}, pages 303--314. ACM, 2018.

\bibitem{tian2018deeptest}
Yuchi Tian, Kexin Pei, Suman Jana, and Baishakhi Ray.
\newblock Deeptest: Automated testing of deep-neural-network-driven autonomous
  cars.
\newblock In {\em Proceedings of the 40th international conference on software
  engineering}, pages 303--314. ACM, 2018.

\bibitem{xiao2018}
Chaowei Xiao, Bo Li, Jun-Yan Zhu, Warren He, Mingyan Liu, and Dawn Song.
\newblock Generating adversarial examples with adversarial networks.
\newblock {\em IJCAI}, 2018.

\bibitem{zhu2016generative}
Jun-Yan Zhu, Philipp Kr{\"a}henb{\"u}hl, Eli Shechtman, and Alexei~A Efros.
\newblock Generative visual manipulation on the natural image manifold.
\newblock In {\em ECCV}, pages 597--613. Springer, 2016.

\end{thebibliography}
}

\end{document}